\documentclass[twoside,dvipsnames]{article}
\usepackage[accepted]{aistats2025}
\usepackage{widetext} 
\usepackage{amsmath}
\usepackage{amssymb}
\usepackage{bigints}
\usepackage{MnSymbol}
\usepackage{graphicx}
\usepackage{float}
\usepackage{svg}
\usepackage{cite} 
\usepackage{enumitem} 
\usepackage{hyperref}

\graphicspath{{./}{./figures/}}

\begin{document}

\newcommand{\mbj}[1]{\textcolor{MidnightBlue}{mbj: #1}}
\newcommand{\vr}[1]{\textcolor{BrickRed}{vr: #1}}
\newcommand{\fl}[1]{\textcolor{blue}{fpl: #1}}
\newcommand{\fsp}[1]{\textcolor{ForestGreen}{fsp: #1}}
\newcommand{\rev}[1]{\textcolor{black}{#1}}

\newcommand*\diff{\mathop{}\!\mathrm{d}}
\newcommand*\sign{\mathop{}\!\text{sign}}
\newcommand*{\rhotrain}{\rho_\mathrm{train}}
\newcommand*{\rhotest}{\rho_\mathrm{test}}
\newcommand*{\rhoopt}{\rho^*}
\newcommand*{\ie}{\textit{i.e.}~}
%

%

\newcommand{\ourtitle}{Class Imbalance in Anomaly Detection:\\Learning from an Exactly Solvable Model} 
\newcommand{\ourrunningtitle}{Class Imbalance in Anomaly Detection} 

\runningtitle{\ourrunningtitle}

\twocolumn[
\aistatstitle{\ourtitle}

\aistatsauthor{F.S. Pezzicoli \And V. Ros \And  F.P. Landes \And M. Baity-Jesi}

\aistatsaddress{TAU team, LISN\\Université Paris-Saclay\\CNRS, Inria\\91405 Orsay, France\\francesco.pezzicoli@\\universite-paris-saclay.fr 
\And LPTMS\\Université Paris Saclay\\CNRS\\91405 Orsay, France\\valentina.ros@cnrs.fr 
\And TAU team, LISN\\Université Paris-Saclay\\CNRS, Inria\\91405 Orsay, France\\francois.landes@inria.fr 
\And SIAM Department\\Eawag (ETH)\\8600 D\"ubendorf, Switzerland\\marco.baityjesi@eawag.ch}
]

\begin{abstract}
Class imbalance (CI) is a longstanding problem in machine learning, slowing down training and reducing performances. Although empirical remedies exist, it is often unclear which ones work best and when, due to the lack of an overarching theory. We address a common case of imbalance, that of anomaly (or outlier) detection. We provide a theoretical framework to analyze, interpret and address CI. It is based on an exact solution of the teacher-student perceptron model, through replica theory. Within this framework, one can distinguish several sources of CI: either intrinsic, train or test imbalance. Our analysis reveals that, depending on the specific problem setting, one source or another might dominate. We further find that the optimal train imbalance is generally different from 50\%, with a non trivial dependence on the intrinsic imbalance, the abundance of data and on the noise in the learning. Moreover, there is a crossover between a small noise training regime where results are independent of the noise level to a high noise regime where performances quickly degrade with noise. Our results challenge some of the conventional wisdom on CI and pave the way for integrated approaches to the topic. 
\end{abstract}




\section{INTRODUCTION}\label{sec:intro}
Supervised learning under class imbalance (CI) is a fundamental challenge in modern machine learning, as many real-world datasets often exhibit varying degrees of imbalance~\cite{yamanishi:00,almeida:11,kyathanahally:21,schur:23}. 
Efforts to mitigate the detrimental effects of class imbalance have led to the development of various approaches, with the machine learning community establishing widely accepted heuristics based on empirical evaluations. 
These approaches can be broadly categorized into three types: those acting on the data distribution~\cite{japkowicz_class_2002,chawla_smote_2002,ando_deep_2017,zhang_knn_2003}, those modifying the loss function~\cite{xie_logit_1989,kini_label-imbalanced_2021,behnia_implicit_2023,thrampoulidis_imbalance_2022,menon_long-tail_2021}, and those biasing the dynamics of the training process~\cite{anand:93,tang:20}. 
However, 
due to the lack of a theoretical framework for the analysis of CI,
it is often unclear which of those methods work best and when or why.
For this reason, recent studies tried to fill this theoretical gap, either by focusing on how imbalance influences the learning dynamics~\cite{ye:21,francazi:23,kunstner:24}, or how it influences the loss landscape~\cite{mignacco2020role,loffredo_restoring_2024,mannelli_unfair_2023}. 

In these works, CI is treated as a single phenomenon, which can be addressed through a single formal approach. We highlight, instead, that one should distinguish between (at least) two types of imbalance. 
What we call \textit{Multiple Groups} imbalance (MGI) involves samples drawn from distinctly different distributions, 
with the imbalance arising either from the sampling process (\textit{e.g.}~the toxicity of certain chemicals is tested more often than others~\cite{schur:23}) or being intrinsic to the data itself (\textit{e.g.}~some species being more common within an ecosystem~\cite{kyathanahally:21}).
In contrast, \textit{Outlier or Anomaly Detection} imbalance (ADI), is generally a binary problem. 
All examples are drawn from the same distribution, and one needs to identify outliers based on an unknown rule (\textit{e.g.}~only some of the components of a power grid will cause a failure, but we do not know what will cause it~\cite{zhang2019anomaly}), with the rule itself determining the imbalance of the data.
In this case, the imbalance is intrinsic to the problem at hand, as anomalies are naturally fewer in number than normal samples. 
As we will see, 
differently from MGI, ADI has an associated intrinsic imbalance scale, which we call $\rho_0$ and which represents the fraction of anomalies. If $\rho_0=0.5$, anomalies and normal samples appear with the same frequency. 

While most of the theoretical literature on class imbalance implicitly treats MGI, we are not aware of theoretical work targeting how ADI affects the loss landscape. This requires a different theoretical setup, 
yields different results, and is the aim of our study.

We study the effects of ADI on the training and test landscape in a paradigmatic analytically tractable model. 
Specifically, we study a modified version of the Teacher-Student (TS) spherical perceptron~\cite{gardner1989three,seung_statistical_1992}, where one can tune the amount of CI, and study its effect on learning.
Studying a tractable model, where the ground truth is known, allows us to disentangle the various reasons why a high performance is reached or not, providing interpretable results.

\paragraph{Contributions.} 

The main contributions of our work are the following :
\begin{enumerate}
    \item By solving the Teacher-Student spherical perceptron in the presence of ADI, we provide an \textbf{interpretable framework to characterize ADI}. This allows us to elucidate the role of three sources of imbalance: intrinsic imbalance $\rho_0$, the train imbalance $\rhotrain$, and the test set imbalance $\rhotest$.  
    As a function of these quantities, we examine how various commonly used performance metrics 
    vary and
    are able to track 
    the quality of the learnt model.

    \item We contribute to challenging the conventional wisdom and common practice that a perfectly re-balanced training set ($\rhotrain=0.5$) is optimal, and instead
    we find that \textbf{the optimal value of $\rhotrain$ is not 0.5}. Factors influencing this value and its relevance include 
    the abundance of data, the amount of noise in the dynamics, and the intrinsic imbalance $\rho_0$. 
    
    \item \textbf{Dynamics with lower noise are less susceptible to CI}. We identify two distinct regions.
    For low noises, the performance is optimal, and largely unaffected by the exact level of noise. 
    For large noises, the performance degrades and is sensitive to additional amounts of noise.
    This effect is related to how well the student can guess the intrinsic $\rho_0$, which is also affected by $\rhotrain$.
\end{enumerate}

\paragraph{Related work.} 
Although recent analytical work covered class imbalance with approaches similar to ours, this does not explicitly distinguish between different types of imbalance, and instead implicitly focuses on MGI. 
These studies assume the presence of two distinct sub-populations in the data distribution and explore the effects of class imbalance from the perspective of the loss landscape, along with potential mitigation strategies. 
\cite{mignacco2020role}
focus on the role of regularization, showing that imbalance impedes achieving Bayes-optimal performances. 
\cite{mannelli_unfair_2023}
focus on fairness implications, studying how imbalance affects the performances across the sub-populations. They also introduce a mitigation strategy based on coupled neural networks trained on subsets of the full training dataset. 
\cite{loffredo_restoring_2024}
investigate the effect of imbalance on various accuracy metrics and in particular show that the AUC score is rather insensitive to imbalance while the Balanced Accuracy is a better suited metrics to study imbalanced problems. They focus also on the effectiveness of re-sampling techniques and prove that mixed strategies of random over-sampling/under-sampling are the most effective.
Their setup explicitly covers MGI, with a learning problem where it is impossible to obtain zero loss.

\begin{figure*}[t]
\centering
\includegraphics[width=.27\textwidth]{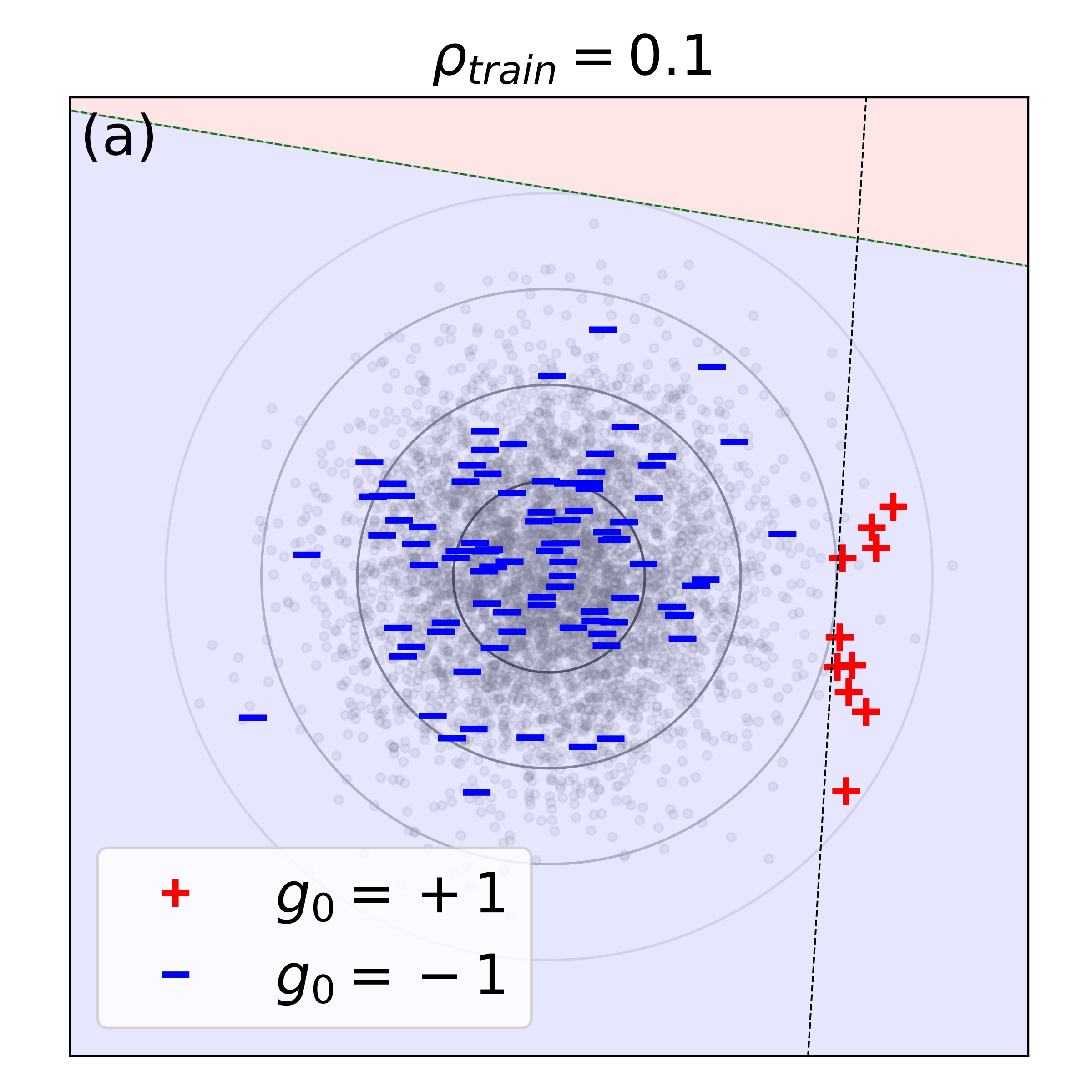}
\includegraphics[width=.27\textwidth]{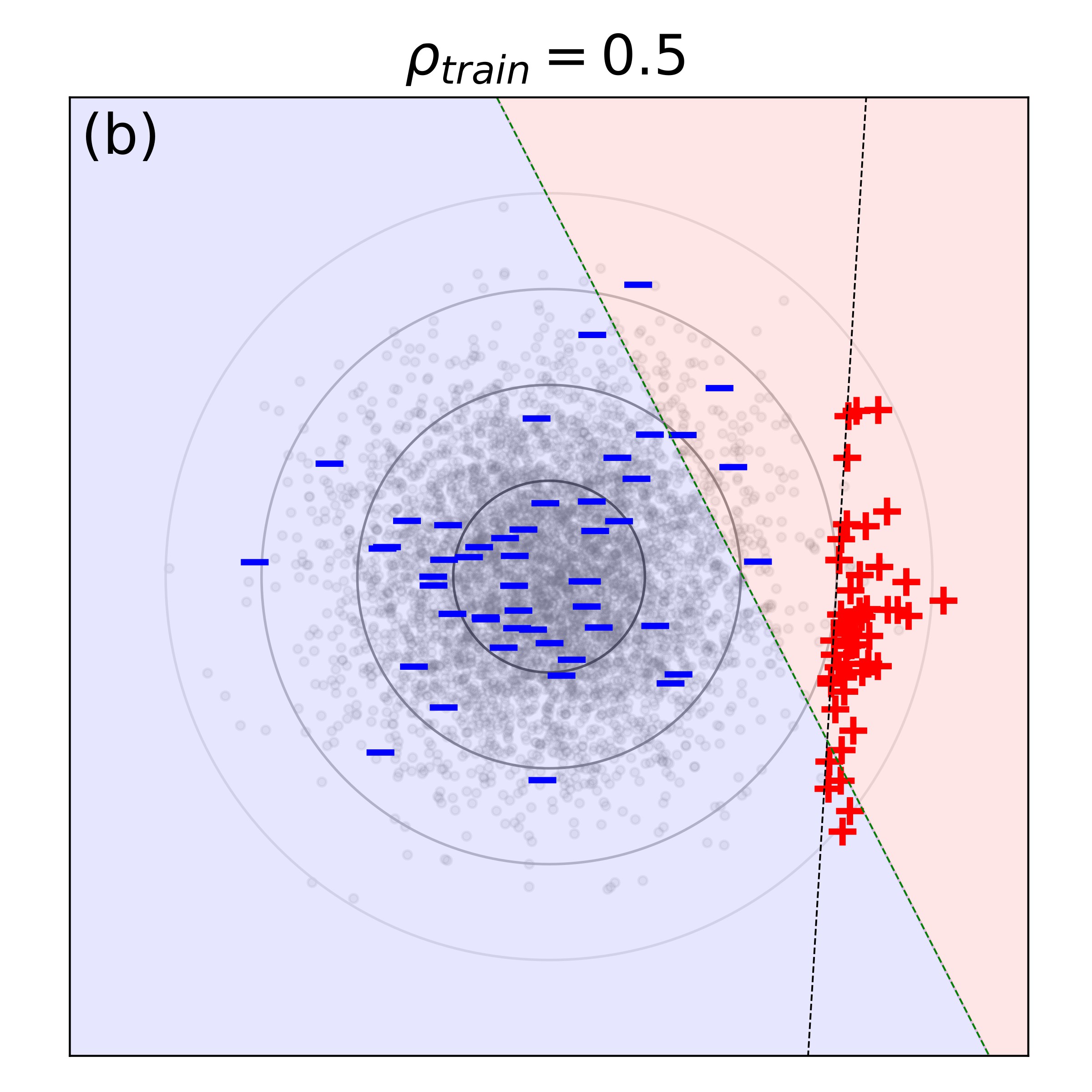}
\includegraphics[width=.27\textwidth]{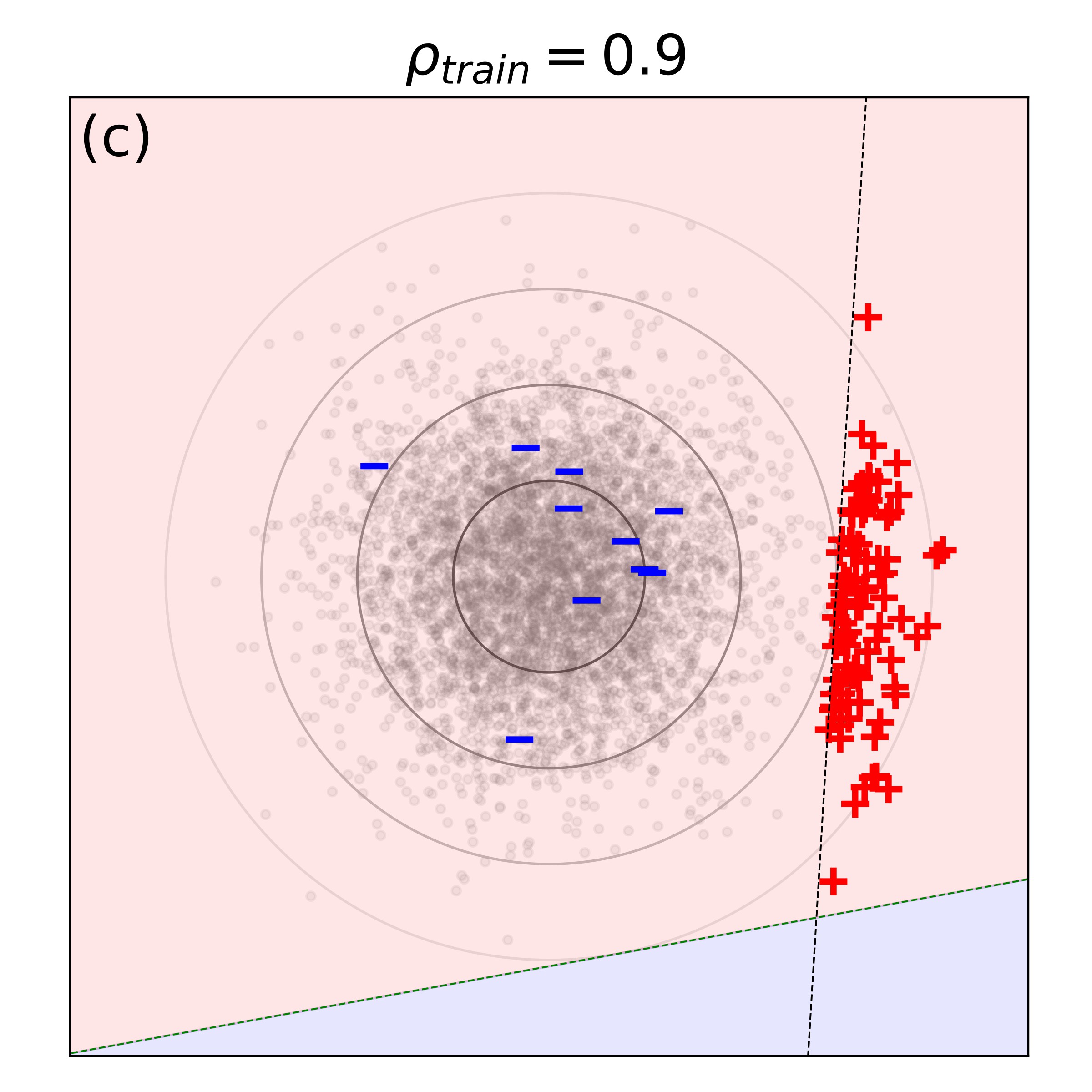}
\caption{\textit{Geometrical interpretation of learning an Anomaly Detection task under class imbalance,} 
with fixed $\rho_0$, and $\rhotrain= 0.1, 0.5, 0.9$.
Normal examples (negative label, $g_0^\ell = -1$) are represented with blue $(\textcolor{blue}{-})$ symbols, anomalies (positive label, $g_0^\ell = +1$) with red $(\textcolor{red}{+})$. Shaded grey points depict the underlying Gaussian data distribution and grey circles locate contours at $1\sigma, 2\sigma$ and $3\sigma$ ($\sigma$ is the standard deviation). The black dashed line represents the teacher hyperplane (decision boundary) which determines the ground-truth labels and the colored regions of the plane depict the classes predicted by the student (the model being trained). The three examples contain the same number of misclassified examples, but the learned model is very different.
When the training set is strongly imbalanced ((a) and (c)) the student has an entropic incentive to learn a strong bias, completely discarding  the alignment with the teacher: learning only a bias that matches the train imbalance is statistically favored due to the large number of possible directions for the student hyperplane that achieve a low error. Learning on a balanced training set (b) forces the student to learn the direction of the teacher because of the higher cost of entirely mis-classifying one of the two classes.
}
\label{fig:introduction}
\end{figure*}

\section{FRAMEWORK}
\subsection{Type of Imbalance}
\label{sec:framework}

\rev{
We now make more precise the nuance between MGI and ADI mentioned in Sec.~\ref{sec:intro}.
\paragraph{Multiple Groups Imbalance}
In MGI, there are $K\geq 2$ separate populations, with distinct feature’s distribution, \ie we have a mixture of densities: $\diff \mu(\mathbf{S}) = \sum_k^K \pi_k \diff \mu(\mathbf{S}|\text{class} =k)$, where $\pi_k$ is the class frequency of class $k$ and $\mathbf{S}$ represents the data. 
Both the distributions $\diff \mu(\mathbf{S}|\text{class} =k)$ and the $\pi_k$ can be different for different $k$. Since the $\pi_k$s are arbitrary, they can be selected in such a way that some classes are more present than others.
\paragraph{Anomaly Detection Imbalance}
In ADI, there is one single population ($K=1$), and the classification problem consists of identifying outliers in the density $\diff \mu(\mathbf{S})$, where the definition of outlier depends on a given criterion/rule. Since this criterion defines both the classes and the frequency of the outliers, the imbalance of the problem (this is what we will call $\rho_0$) cannot be changed once the classes are defined. In this paper, our whole theoretical setup will be in the ADI setting.
}
\subsection{Model}
\label{sec:model}
We consider the widely-studied Teacher-Student Spherical Perceptron~\cite{gardner1989three,gyorgyi_first-order_1990,seung_statistical_1992,fontanari_statistical_1993,nishimori2001statistical}. In this set-up, {diagrammatically represented in Fig.~\ref{fig:sketch_TS} in App.\ref{app:TS},} a \textit{teacher} perceptron with a planted weight configuration assigns a label to each sample, while a \textit{student} perceptron learns to mimic the teacher by adjusting its weights through Empirical Risk Minimization on the samples in the training set. 
\rev{We motivate these modeling hypotheses through empirical observations
in App.~\ref{app:dataDistributionisADI}, by studying the distribution of input features in a realistic anomaly detection task, namely for the BTAD dataset \cite{mishra2021}. Those observations justify adopting 
the ADI setup (rather than MGI).} 

\paragraph{Problem Setting.}
Given an input sample, $\mathbf{S}^\ell\in\mathbb{R}^N$ ($\ell$ is the sample index and $N$ the number of features), the Spherical Perceptron model related to the student assigns it a label, $g^\ell$, through the relation
\begin{equation}
    g^\ell = g \left(  \frac{\mathbf{w}\cdot \mathbf{S}^\ell}{\sqrt{N}} + b  \right)\,,
    \label{eq:gell}
\end{equation}
where $\mathbf{w}=(w_1,\dots,w_N)\in\mathbb{R}^N$ are the model weights, subject to the constraint $\mathbf{w}^T \cdot \mathbf{w} = N$, $b\in \mathbb{R}$.
{This constraint is akin to a regularization, and ensures that each weight is of order $1$.}
The activation function is usually chosen as $g(x) = \sign{(x)}$. 
Ground-truth labels $g_0^\ell$ are obtained through the teacher assignment: 
\begin{equation}
    g_0^\ell = g \left(  \frac{\mathbf{w}^0\cdot \mathbf{S}^\ell}{\sqrt{N}} + b_0  \right)\,.
\end{equation}
Although teacher and student have the same architecture, the teacher's parameters $\mathbf{w}^0$ and $b_0$ are fixed, while the student parameters $\mathbf{w}$ and $b$ are learned through the minimization of the loss
\begin{equation}
    \mathcal{E}(\mathbf{w},b) = \sum_{\ell=1}^{P} \epsilon \left(  g \left(  \frac{\mathbf{w}\cdot \mathbf{S}^\ell}{\sqrt{N}} + b  \right), g_0^\ell\right)\,,
\label{eq:loss}
\end{equation}
where  $P$ is the number of training samples.
We use a square loss, $\epsilon(x,y) = \frac{1}{2}(x-y)^2$.
The shape of the loss matters little, since in practice errors have cost $+2$ and correct predictions have cost $0$.

\paragraph{Modeling Class Imbalance.}
\label{subsec:modeling_class_imbalance}
We model the ADI by introducing an imbalance parameter 
$\rhotrain$, 
that fixes the ratio between samples in the majority and minority classes in the training set.

The set of all data-points observed during training is denoted as $\{\mathbf{S}^\ell\}_{\ell=1}^P$ with $\mathbf{S}^\ell = (S_1^\ell,\dots,S_N^\ell) \in \mathbb{R}^N$ and $P= N \alpha$.  We fix the ratio $\alpha${, which represents the abundance of data,} 
to be finite.
{This is a classical choice in this setup, which ensures that the problem is non trivial, \textit{i.e.}~it is neither overconstrained nor underconstrained.}
Samples are i.i.d., and for each sample $\mathbf{S}$ the components are distributed according to: $S_i \sim DS_i = \frac{\diff S_i}{\sqrt{2\pi}} e^{-S_i^2/2}$. We will use the shorthand notation $D\mathbf{S} = \prod_{i=1}^N DS_i$ to denote the measure of probability of the single sample. 

In order to have a skewed distribution of samples with a number $N\alpha\rhotrain$ of positive (anomalous) samples ($g_0^\ell = +1$) and $N\alpha(1-\rhotrain)$ of negative (normal) samples ($g_0^\ell = -1$), we define the training set measure as:

\begin{widetext}
\begin{equation}
    \diff \mu_{\text{train}} (\{\mathbf{S}^\ell\}) = \frac{1}{c_+^{N\alpha\rhotrain}}\left(\prod_{\ell=1}^{\alpha N \rhotrain} D\mathbf{S}^\ell \Theta\left(  \frac{\mathbf{w}^0\cdot \mathbf{S}^\ell}{\sqrt{N}} + b_0  \right) \right) 
    \frac{1}{c_-^{N\alpha(1-\rhotrain)}}\left(\prod_{\ell'=\alpha N \rhotrain + 1}^{\alpha N} D\mathbf{S}^{\ell'} \Theta\left(  -\frac{\mathbf{w}^0\cdot \mathbf{S}^{\ell'}}{\sqrt{N}} - b_0  \right) \right)\,.
    \label{eq:mu_train}
\end{equation}
\end{widetext}
The notation $\diff \mu (\{\mathbf{S}^\ell\})$ is shorthand for $\diff \mu (\mathbf{S}^1,\dots,\mathbf{S}^{N\alpha})$ and denotes the measure of probability over all the training samples.
We use the Heaviside ($\Theta$) function to select the samples according to the relevant output sign of the Teacher Perceptron. The constants $c_+ = 1/2\,\text{Erfc}(-b_0/\sqrt{2})$, and $c_- = 1 - c_+$, represent respectively the normalization constant for positive and negative samples.
Note that $ \diff \mu_{\text{train}} $ is in general not a Gaussian measure, since in the direction of $\mathbf{w}^0$, it is a piecewise Gaussian with normalization factors which depend on $\rhotrain$: see Fig.~\ref{fig:introduction} for a sketch in two dimensions.

\begin{figure*}[t]
\includegraphics[width=.245\textwidth]{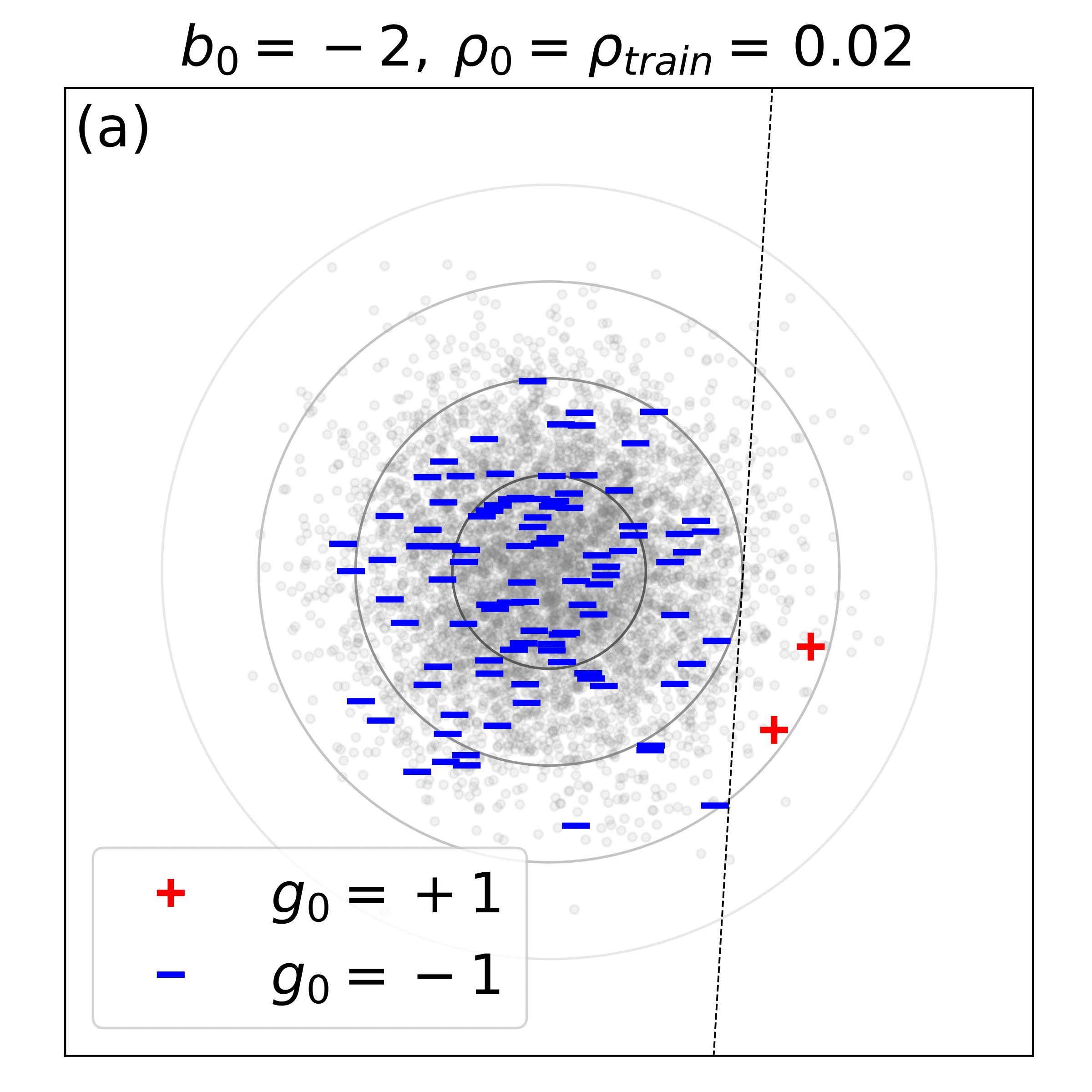}
\includegraphics[width=.245\textwidth]{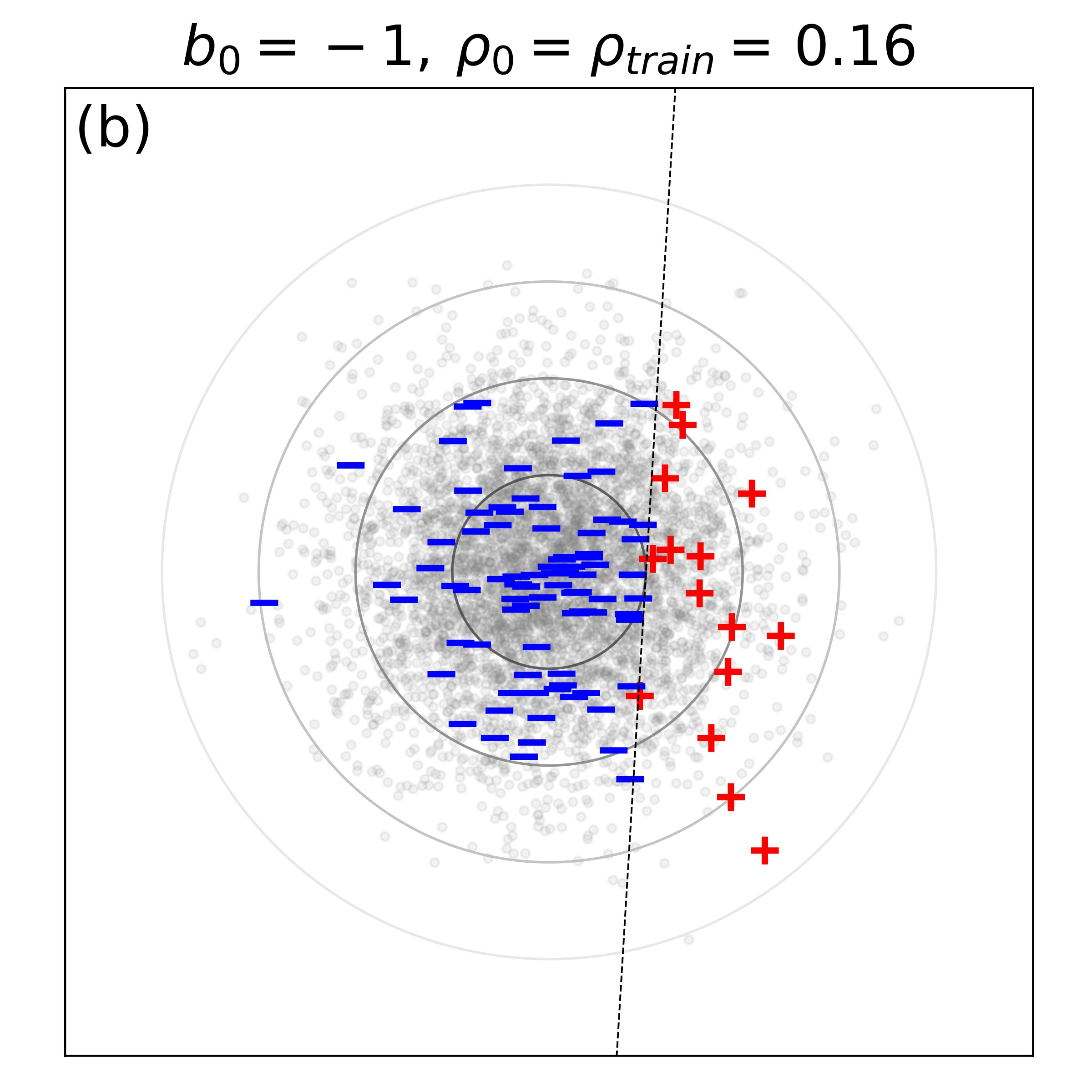}
\includegraphics[width=.245\textwidth]{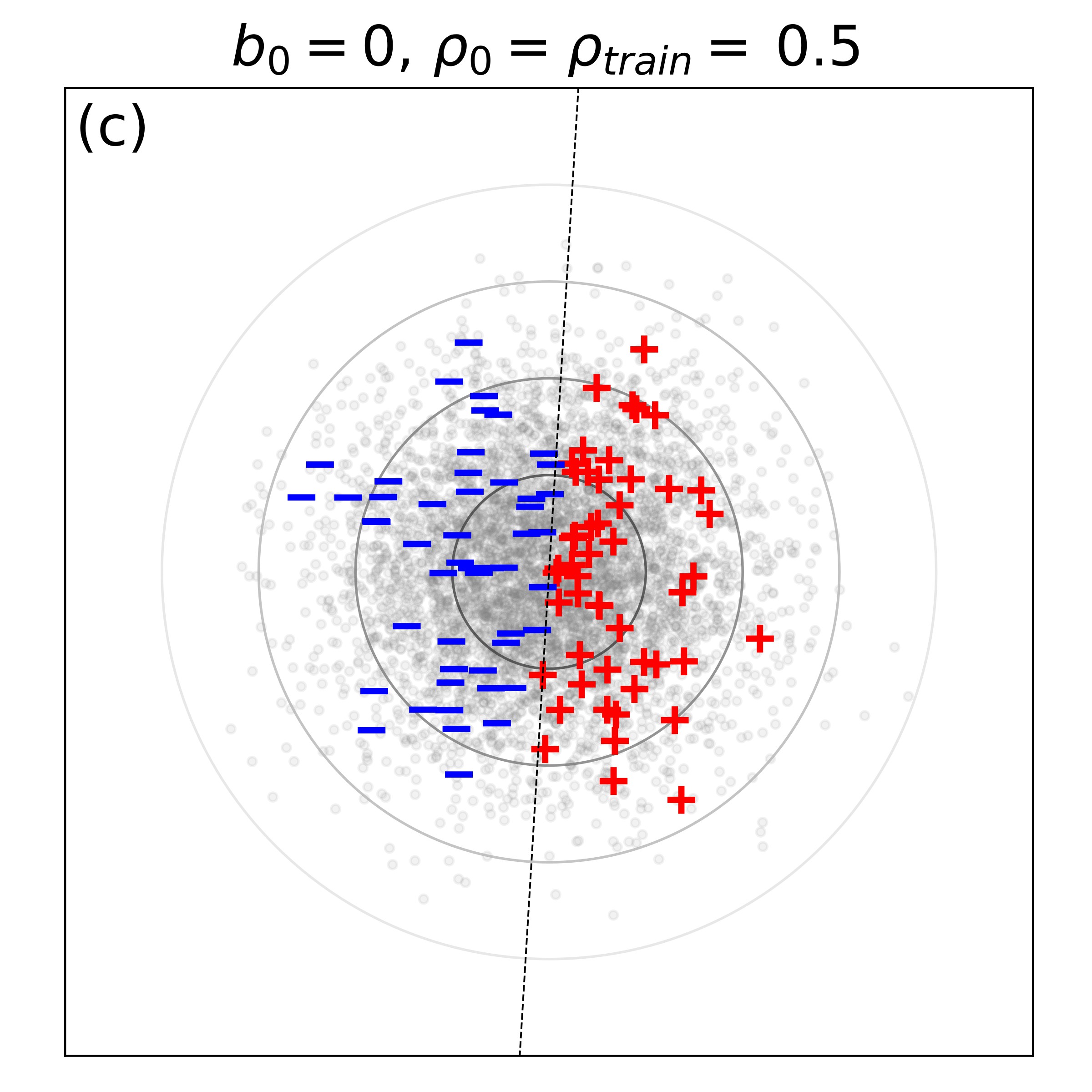}
\includegraphics[width=.245\textwidth]{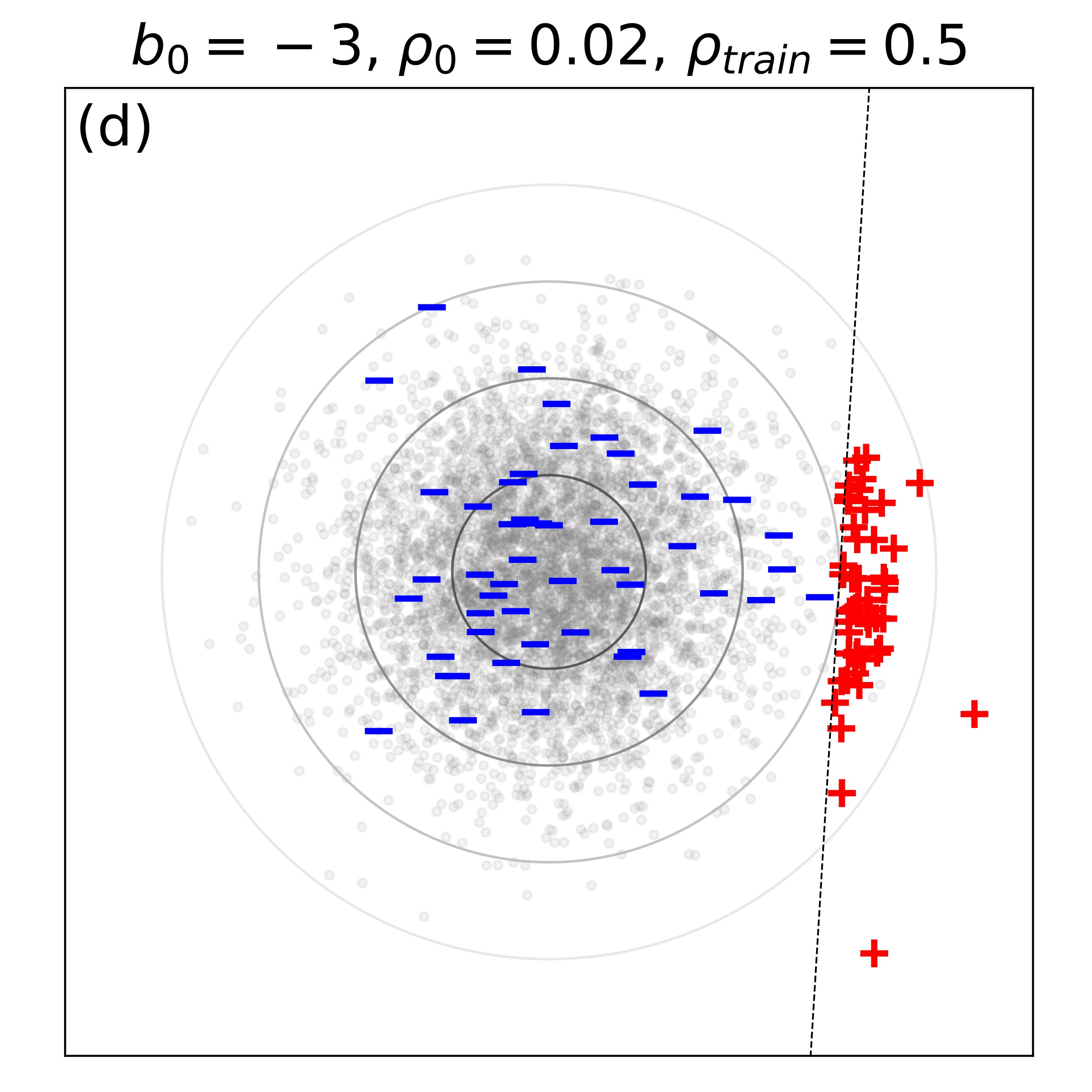}
\caption{\textbf{Plots (a,b,c):} \textit{Intrinsic Imbalance}. Two-dimensional sketches showing the effect of the teacher bias $b_0$ on the intrinsic imbalance $\rho_0$. 
The symbols in the figures are as in Fig.~\ref{fig:introduction}. All the examples are extracted from the underlying Gaussian distribution and no specific imbalance ratio is imposed externally. Increasing the magnitude of the teacher bias $b_0$ ((c) $\to$ (a)) translates the teacher's hyperplane (black dashed line) away from the origin and makes anomalies more rare, as the Gaussian tail is cut further away.
\textbf{Plots (c,d):} \textit{Informative samples}. Two cases are compared where the train imbalance is fixed to $\rhotrain=0.5$ 
while the teacher biased is varied.
As $|b_0|$ grows, anomalies become more and more concentrated around the teacher's hyperplane, becoming more informative about the teacher's direction.}
\label{fig:features_data}
\end{figure*}

\paragraph{Intrinsic, Train and Test Imbalance.} Here, we elaborate on the definition and roles of the imbalance ratios $\rho_0$, $\rhotrain$ and $\rhotest$.

The introduction of $\rhotrain$ in Eq.~(\ref{eq:mu_train}) allows to control the amount of imbalance in the \textbf{training set}. This parameter is externally imposed and enables us to explore scenarios where the model is trained with varying levels of imbalance.

Instead, 
when samples are  extracted from the Gaussian measure $\diff \mu_{\text{pop}} (\{\mathbf{S}^\ell\}) \propto \prod_{\ell=1}^{\alpha N } D\mathbf{S}^\ell$, an imbalance between classes arises naturally due to the bias parameter $b_0$. It can be computed as:
\begin{equation}
    \rho_0 (b_0) = P\left(\frac{\mathbf{w}_0 \cdot \mathbf{S}}{\sqrt{N}}+b_0>0\right) = \frac{1}{2}\text{Erfc}\left(-\frac{b_0}{\sqrt{2}}\right)=c_+    \,.
\end{equation}
{As soon as $b_0\neq 0$, we have $\rho_0 \neq 0.5$, \textit{i.e.}~there is intrinsic imbalance.}
We refer to this $\rho_0$ as \textbf{Intrinsic Imbalance} as it measures the intrinsic imbalance present in the data generation process. 
It describes the rarity of observing anomalies and in our case, it depends solely on the teacher's bias.
It can easily be visualized geometrically: when dealing with the population, all the samples are drawn from a Gaussian distribution centered at the origin, while varying $b_0$ amounts to translating the Teacher's hyperplane away from the origin of the $N$-dimensional space. Thus, one of the two classes lies on the tail of the distribution and is less represented. This is depicted in 2D sketches of Fig.~\ref{fig:features_data}(a,b,c), where a decreasing $b_0$ (in magnitude) results in a less biased teacher and a less imbalanced population.

Since the population distribution is rarely available in practice, a \textbf{test set} consisting of samples not observed during training is introduced in standard machine learning settings to test the performance of the trained model. We define the test distribution as 
\begin{align}
    \diff \mu_{\text{test}} (\mathbf{S}) = \frac{\rhotest}{c_+}
    &\Theta\left(  \frac{\mathbf{w}^0\cdot \mathbf{S}}{\sqrt{N}} + b_0  \right) D\mathbf{S} +\nonumber
    \\
    +\frac{1-\rhotest}{c_-}
    &\Theta\left(  -\frac{\mathbf{w}^0\cdot \mathbf{S}}{\sqrt{N}} - b_0  \right) D\mathbf{S}\,,
\end{align}
where $\rhotest$ measures the probability of having an anomaly in the test set. Common choices are $\rhotest = 0.5$ (balanced test set) or $\rhotest = \rho_0 (b_0)$ (test set that reflects the intrinsic imbalance). As we will discuss below, while some performance metrics do not explicitly depend on  $\rhotest$, others do. The choice of the test imbalance is as important as the train imbalance, and can lead to misleading results if not properly considered, by inducing choices of $\rhotrain$ which do not allow to properly reconstruct the teacher model.

\paragraph{Informative samples.}
The teacher bias $b_0$ also determines how informative the two classes are.
When $b_0\neq 0$, the anomalous samples 
concentrate on the hyperplane of the teacher
(in our convention $b_0<0$ and it is the class $+1$ which is the anomaly and concentrates), while the samples of the other class have a lower probability to lie close to this boundary. 
In particular, some of the anomaly-class points will be close to the hyperplane but far from the projection of the origin onto the hyperplane, and these are 
expected to be 
more informative about its location.
Fig.~\ref{fig:features_data}(c,d) shows this effect through two-dimensional sketches. 
In App.~\ref{app:informative} we demonstrate this quantitatively, by calculating the density of each class close to the teacher's hyperplane. As $|b_0|$ grows, points close to this boundary are predominantly from the anomalous class.
This effect is dominant when the student has perfect information about the teacher's bias, as discussed in App.~\ref{app:fixedb}.

\section{THEORETICAL ANALYSIS}
\label{sec:theoretical_analysis}
\subsection{Statistical Mechanics Approach}
We are interested in finding the configurations of the student ($\mathbf{w},b$) that minimize the Loss in Eq.~\eqref{eq:loss}
and in computing the properties of these configurations over the distribution of input samples. 

Statistical Mechanics (SM) comes in handy for this problem, since it allows us to exactly calculate properties of the typical solutions of our model, in the large-model limit, $N\to \infty$. This is the relevant limit for the study of our system, provided that the number of constraints is proportional to the number of weights:  $\alpha=P/N$ is finite~\cite{roberts2022principles}. 
Since the derivation of our results is long and technical, we here focus on the hypotheses and on the interpretation of the results, and provide the full calculation in App.~\ref{app:sec:replica_calculations}, where we calculate all the quantities described in the main paper, using Replica Theory (see \textit{e.g.}~\cite{altieri:24} for a thoroughly referenced review). 

SM assumes that the model configurations are sampled through Langevin dynamics at a temperature $T$, which roughly corresponds to the ratio between the learning rate and the batch size~\cite{jastrzkebski:17}. 
Analyzing the system at 
$T=0$ returns the properties of the absolute minima of the loss; 
doing it for $T > 0$ identifies the typical configurations that are reached with a certain level of noise in the dynamics. 

The main quantity is the free energy $F$, which is the 
negative 
log-probability of a weight configuration, so the most probable configurations are those that minimize $F$. In the $N\to\infty$ limit, the original high dimensional optimization problem reduces to a system of coupled deterministic equations for some order parameters, $(R,\hat{R},q,\hat{q},b)$, which must be solved through recursive iteration (see App.~\ref{app:sec:replica_calculations}). 
These parameters characterize the typical configurations that are reached by the Student at the end of the training, \textit{i.e.}~when training is long enough to reach an equilibrium state. 
This means that, once the values of the order parameters are given, any other quantity derives deterministically from them. 
The order parameters $R$ and $b$ have a crucial interpretation; $R$ is the Teacher-Student overlap and it's defined as
$R = \lim_{N \to \infty}\frac{\mathbf{w}\cdot\mathbf{w}^0}{N}$.
It measures the typical alignment between the teacher's and the student's hyperplane. 
Instead, $b$ represents the typical bias learned by the student, and should be compared to the teacher's value, $b_0$. 

\paragraph{Calculating Metrics.}
For any given choice of the hyperparameters $b_0,\rhotrain,T$ and $\alpha$ (these are also called control parameters, although $b_0,\alpha$ are intrinsic to the ML task and cannot be controlled in practical settings), we can solve the set of self-consistent equations (Eqs.~(\ref{eq:saddle1}--\ref{eq:saddle5}) in App.~\ref{app:sec:replica_calculations}) numerically, to obtain the order parameters, with particular interest in $(R,b)$.

The generalization error and the other performance metrics on the test-set are all derived from the order parameters. For any metric $M(\mathbf{w},b;\mathbf{S})$, its generalization value is:
\begin{align}
    M_g = \llangle  \llangle \mathbb{E}_T[M(\mathbf{w},b;\mathbf{S})] \rrangle_{\mu_{\text{train}}}\rrangle_{\mu_{\text{test}}}\,,
\end{align}
where $\mathbb{E}_T[\dots]$ denotes the average over realizations of the thermal noise, and $\llangle\ldots\rrangle_\mu$ the average over 
a chosen dataset 
measure $\mu$.
The idea behind the computation of generalization metrics is to add one sample that was not observed during training and evaluate the performance of the trained student on it. 
In App.~\ref{app:sec:train_and_generalization_metrics},
we provide the derivation of the expressions of all the quantities shown in the paper.

The performance metrics we are interested in are: Recall ($r$), Specificity ($s$), Accuracy ($a$), Balanced Accuracy ($a_\mathrm{bal}$), Positive Predicted Value or Precision ($p$), F1-Score ($F_1$), generalization error ($\epsilon_g$): 
\begin{align}
    \label{eq:r}
    r &= \frac{\# \text{True Positives}}{\# \text{Teacher Positives}}\,,\\[1ex]
    \label{eq:s}
    s &= \frac{\# \text{True Negatives}}{\# \text{Teacher Negatives}}\,,\\[1ex]
    \label{eq:a}
    a &= \rhotest r + (1 - \rhotest) s\,,\\[1ex]
    \label{eq:ba}
    a_\mathrm{bal} &= \frac{(r + s)}{2}\,,\\
    \label{eq:p}
    p &= \frac{r}{r + \frac{1-\rhotest}{ \rhotest} (1-s)}\,,\\
    \label{eq:f1}
    F_1 &= 2\cdot \frac{p r}{p + r}\,,\\
    \label{eq:err}
    \epsilon_g &= 1-a\,.
\end{align}

\subsection{Theoretical Results}
\label{subsec:th_results}
\begin{figure}[!ht]
\includegraphics[width=0.45\textwidth]{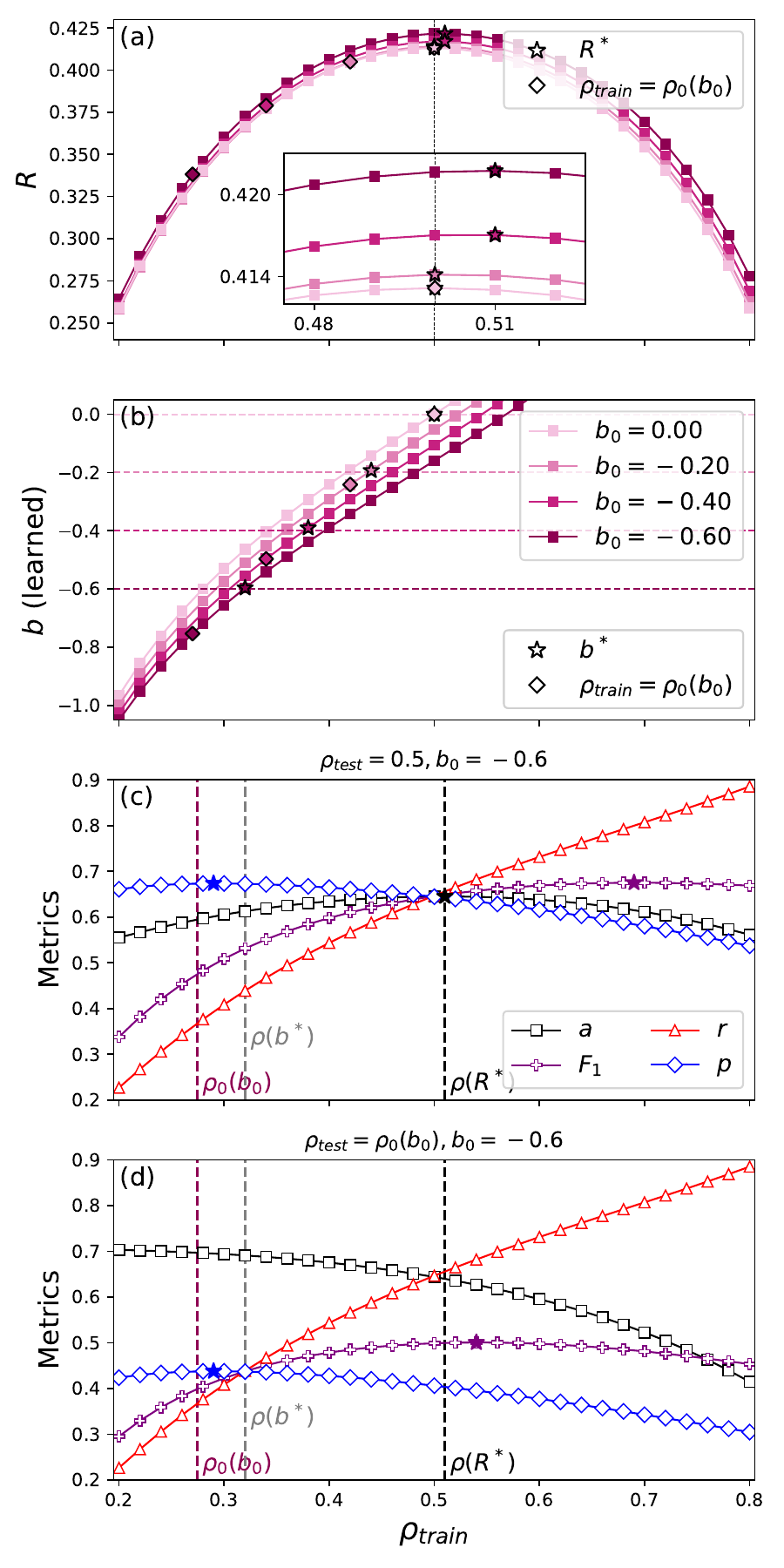}
\caption{
\textit{Performance as function of $\rhotrain$.}
Analytical results as a function of $\rhotrain$, for $\alpha=1.1$ and $T=0.5$. 
\textbf{(a):} Student overlap $R$, for $b_0=0,-0.2,-0.4,-0.6$ ($\rho_0=0.5, 0.42, 0.34, 0.27$). Stars indicate maximal overlap point, diamonds indicate $\rhotrain=\rho_0$. Vertical line indicates $\rhotrain=0.5$. Inset is a zoom. 
\textbf{(b):} As in (a), but for the student bias $b$. Horizontal lines indicate $b_0$. Stars now indicate points where bias is learned perfectly ($b=b_0$), diamonds indicate $\rhotrain=\rho_0$.
\textbf{(c):} Metrics 
for $b_0=-0.6$, $\rhotest=0.5$. Stars indicate the peak of each curve. Vertical lines indicate $\rho_0$ (imbalance $\rho(b^*)$ at which the bias is optimal), and that at which the overlap is optimal, $\rho(R^*)$.
\textbf{(d):} Same as (c), but for $\rhotest=\rho_0(b_0)$. 
}
\label{fig:theoretical_results_vs_rho}
\end{figure}

\paragraph{Good and bad models, Energy-Entropy interplay.}
We now investigate the influence of $\rhotrain$ on the learned model. 
Figure~\ref{fig:theoretical_results_vs_rho}(a,b), reports the solution of the self-consistent equations
\footnote{Although our results are analytical, each point in the plot is a new solution of the system of self-consistent equations (Eqs.~(\ref{eq:saddle1}--\ref{eq:saddle5}) in the appendix). For this reason, our analytical  results are provided as points instead of a continuous curve.}
for the overlap $R$ and the learned bias $b$ as a function of $\rhotrain$, for multiple choices of the teacher bias $b_0$.
Learning under strong imbalance leads to a model that has a strong bias and low alignment with the teacher, this is a \textit{bad} model since it is not able to reproduce $b_0$ and $\mathbf{w}_0$ correctly. 
Meanwhile, learning on a more balanced training set, leads to a \textit{good} model that is able to better reproduce the teacher's labeling rule, learning a good overlap $R$ and not being overly biased. 
This phenomenon can be geometrically interpreted (Fig.~\ref{fig:introduction}(a,c)): since the loss counts the number of misclassified examples, a dummy model that has a strong bias and always predicts the majority class pays a small price in terms of loss (for small $\rhotrain$). However, such a model is statistically favored with respect to a model with $b=b_0$, since with $|b|\gg|b_0|$, there is a very large number of weight configurations $\mathbf{w}$ that allow for the same small training error, while with $b=b_0$ the number of weight configurations giving a small error is much lower. This is an example of what is called energy-entropy interplay~\cite{carbone:20}: solutions with large $b$ have an entropic advantage (there are more of them), at the expense of a few misclassifications (what is sometimes called an \textit{energetic cost}). As $\rhotrain$ becomes more balanced, the energetic cost increases, eventually overcoming the entropic advantage.
We also note that, while the situations in Figs.~\ref{fig:introduction}(a,c) are similar with what regards the training errors, they are of course very dissimilar when one looks into the generalization error (accuracy curve in Fig.~\ref{fig:theoretical_results_vs_rho}(d)).\\
Finally, we highlight that training at $\rhotrain=\rho_0$ is always suboptimal, both in terms of $R$ and of $b$.

\paragraph{Performance metrics, which one?}
When testing a trained model, it is common practice to build a balanced test set with samples not observed during training, \textit{i.e.}, $\rhotest = 0.5$. Another practice is to leave the distribution untouched, $\rhotest = \rho_0(b_0)$. How are different performance metrics affected by the test imbalance, and which metric is best able to identify a \textit{good} model in terms of $R$ and $b$? Figure~\ref{fig:theoretical_results_vs_rho}(c,d) reports different performance metrics for models trained with varying $\rhotrain$ and tested with $\rhotest=0.5$ and $\rhotest=\rho_0$. 
\\
The recall is trivially maximized for $\rhotrain=1$, since this generates a dummy model which identifies anything as an anomaly. The sensitivity $s$ has the opposite trend (Fig.~\ref{app:fig:theoretical_results_vs_rho_all_metrics} in App.~\ref{app:sec:train_and_generalization_metrics}), being trivially maximized for $\rhotrain=0$.
\\
The balanced accuracy (shown implicitly as accuracy in Fig.~\ref{fig:theoretical_results_vs_rho}(c), since $a_\mathrm{bal}$ coincides with $a$ when $\rhotest=0.5$, Eq.~\eqref{eq:ba}) is the quantity that best reproduces $R(\rhotrain)$, and shares with $R(\rhotrain)$ the feature of not depending on $\rhotest$.
\\
For $\rhotest=\rho_0$, the accuracy $a$ 
is maximized at small $\rhotrain$, because this generates a model which always guesses the majority class. This can be understood from Eq.~\eqref{eq:a}: if $\rhotest \sim 0$ then the contribution of $s$ dominates.
\\
The trend of the precision $p$ (confirmed in  App.~\ref{app:rhostar}) seems independent of $\rhotest$, though its specific value is.

The precision $p$ peaks between $\rho_0$ and $\rho(b^*)$, making it the best candidate to identify $b_0$. Although the precision is a metric that is notoriously dependent on $\rhotest$~\cite{burkard:25}, the position of the peak does not seem to have a strong dependence on $\rhotest$.
While, quantitatively, this observation may vary with $\alpha$ and $T$, we show in App.~\ref{app:rhostar} that $p$ is still the metric that peaks closest to $\rho_0(b_0)$.
\\
Finally, the $F_1$ score, when calculated with $\rhotest=0.5$, peaks at a value representing a low $R$ and a bias which is underestimated compared to $b_0$.
\\
Comparing panels (c) and (d) we note that some metrics ($a$ or $F_1$ score) do not peak at the same location, highlighting the relevance of the choice of $\rhotest$. 
\\
In summary, 
no known metric perfectly matches the optimal overlap ($R^*$) nor the optimal bias ($b^*$) over the whole space of control parameters ($b_0, \alpha, T)$ but
we identify $a_\mathrm{bal}$ as the most suitable metric to identify the overlap, due to the qualitative agreement with $R$ and the independence from $\rhotest$, and 
$p$ as the most suited to track the optimal $b$.


\begin{figure}[t]
\centering
\includegraphics[width=\columnwidth]{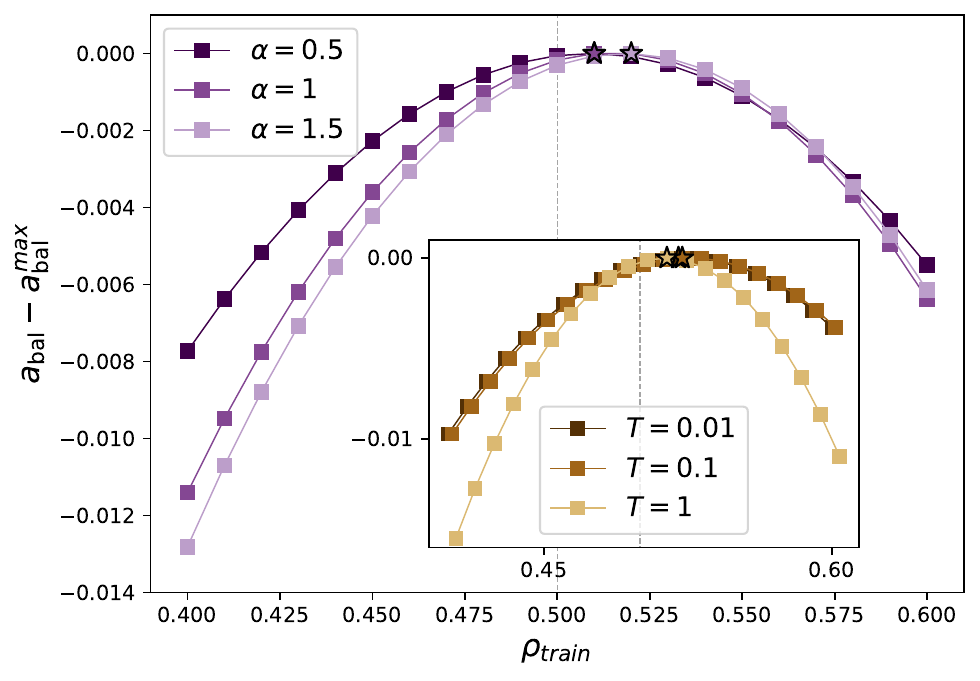}
\caption{
\textit{Dependence on $\rhotrain$ for different $\alpha$ or $T$.}
The optimal balanced accuracy ($a_\mathrm{bal}$).
We plot $a_\mathrm{bal}$ as a function of $\rhotrain$, shifted so that all the curves peak at 0. The vertical dotted lines indicate $\rhotrain=0.5$.
\textbf{Main:} study at $b_0 = -1$ and $T=0.5$. Varying $\alpha$ changes the position of the peak, as well as how fast the performance decreases when leaving the peak. 
\textbf{Inset:} same $b_0$ and fixed $\alpha=1.1$, varying $T$. The cases $T=0.01$ and $T=0.1$ are almost impossible to distinguish because they both correspond to the low-temperature region (Fig.~\ref{fig:theoretical_results_vs_T}). For high $T$ the curvature is larger.
}
\label{fig:theoretical_results_best_tilt}
\end{figure}

\paragraph{Optimal train imbalance.}
We already noted that 
training at $\rhotrain=\rho_0$ never gives the best model (in terms of best $R$ nor $b$). 
We now turn to $\rhotrain=0.5$, which is commonly believed to lead to optimal generalization performances, and the most common choice in CI reweighting/resampling schemes. We challenge this assumption, showing that the optimal train imbalance, $\rhotrain(R^*)=\mathrm{argmax}_{\rhotrain}(R)$, is different from $0.5$. This is true for the overlap (Fig.~\ref{fig:theoretical_results_vs_rho}(a)--inset), and it is also true for the best proxy of the overlap, $a_\mathrm{bal}$.
Fig.~\ref{fig:theoretical_results_best_tilt} shows that, when $b_0<0$ and $\alpha=1.1$, $a_\mathrm{bal}$ peaks at $\rhotrain>0.5$,
\ie when there are slightly more of the anomalous examples. In App.~\ref{app:rhostar} we investigate extreme values of $\rho_0(b_0)$ (down to $\rho_0=3\cdot10^{-7}$) and confirm these findings.
This is consistent with previous empirical observations on SVMs and Random Forests, which found that $\rhotrain=0.5$ is not the optimal training ratio~\cite{kamalov_partial_2022}.

We also look into the influence of the degree of 
data abundance 
($\alpha$) and the amount of noise in the dynamics ($T$) on the value of $\rhotrain(R^*)$: for the values considered in Fig.~\ref{fig:theoretical_results_best_tilt},
increasing $\alpha$ and decreasing $T$ make the curves more tilted, shifting the optimal train imbalance, and the curves more peaked, thus increasing the penalty for choosing $\rhotrain\neq\rhotrain(R^*)$. 


In App.~\ref{app:rhostar} and in particular in Fig.~\ref{app:fig:rhostar_vs_alpha} we further investigate these effects, showing that $\rhotrain(R^*)$ depends on $\alpha, T$ and $b_0$;  that the effect is non-monotonic; and it can shift $\rhotrain(R^*)$ both to values $>0.5$ (as in Fig.~\ref{fig:theoretical_results_best_tilt}) and $<0.5$.
\\
We argue that this is the result of two competing effects.
On one side, as depicted in Fig.~\ref{fig:features_data}(d), minority class examples are more informative, so it is more convenient to train with more of those (\ie increase $\rhotrain$). On the other side, if $|b_0|$ is large enough, there is a large region $\mathcal{R}$ between typical negatives and (informative) positives that is empty of points, thus allowing for many possible hyperplane directions $\mathbf{w}$ that separate the training set. This means that a large fraction of student models with $|b|<|b_0|$ will result in a small error. Since there are many more weight configurations allowing for $|b|<|b_0|$ than weight configurations allowing $b=b_0$, configurations with a wrong $b$ and $\mathbf{w}$ are entropically favored. One way to decrease this entropic contribution is to fill the region $\mathcal{R}$ with negatives, \ie increase the proportion of negatives (decrease $\rhotrain$).

\begin{figure}[t]
\centering
\includegraphics[width=0.49\columnwidth]{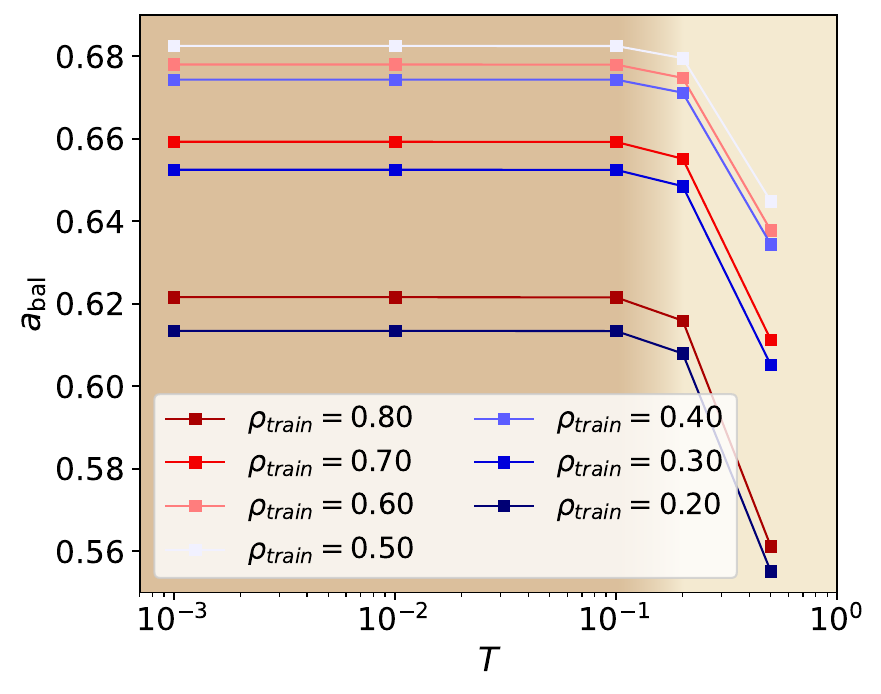}
\includegraphics[width=0.49\columnwidth]{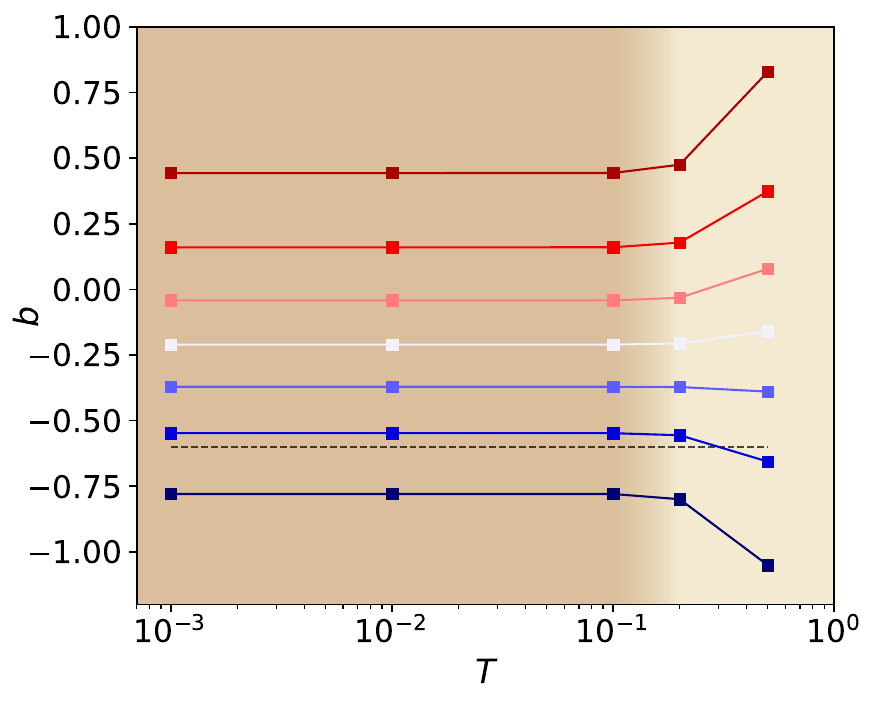}
\caption{
\textit{Performance as function of $T$.}
\textbf{Left:} 
Balanced accuracy as a function of temperature $T$, for $\alpha=1.1$.
The teacher bias is $b_0=-0.6$ (dotted horizontal line in the inset, $\rho_0=0.27$). 
\textbf{Right:}
Same, for the learned bias $b$.
}
\label{fig:theoretical_results_vs_T}
\end{figure}

\paragraph{Interplay between noise and CI.} We investigate the impact of the amount of noise in the dynamics ($T$) on the quality of the learned model. Fig.~\ref{fig:theoretical_results_vs_T} shows a crossover around a temperature $T^*$. Below $T^*$ the performance is optimal, and largely unaffected by the noise level. Above $T^*$, the performance degrades and becomes sensitive to any additional amount of noise. By comparing with Fig.~\ref{fig:theoretical_results_best_tilt}--inset, we see that the low performance in the high-noise regime is connected to its lower tolerance to non-optimal values of $\rhotrain$ (more peaked shape).

A similar effect has already been observed in the teacher-student perceptron, from a dynamical perspective, by studying the regimes of Stochastic Gradient Descent (SGD) as shown in \cite{sclocchi_different_2024}. They identify a Gradient Descent-like regime with low noise and optimal performances and a noise-dominated regime where performance deteriorates as noise increases. Here, we elucidate the interplay between the noise level and the train imbalance. In particular, we observe that $\rhotrain$ determines the evolution of the student's bias with $T$: increasing $T$ favors the entropic contribution discussed 
earlier in this section (seen in Fig.~\ref{fig:introduction}) 
and this pushes towards overly-biased models, because there are more of such solution: \ie dummy models that classify most points as positives (for large $\rhotrain$) or as negatives (for smaller $\rhotrain$), despite making a number of (training) errors.
This is a typical energy-entropy interplay.
To summarize, it is the difficulty to guess the correct bias (or guess the corresponding $\rho_0$) that explains the noise dependence.


\subsection{Experiments}
\label{sec:experiments}
Most of the findings discussed above emerge from the study of the static properties of the loss landscape. The dynamics (which in our calculations is Langevin) enters the discussion only through the parameter $T$, which measures the amount of stochastic uncorrelated noise. Since in practical experiments the dynamics used is SGD, our experiments are with SGD dynamics too.
In App.~\ref{app:experiments} we provide empirical evidence which is consistent with our analytical findings, in three cases: (i) in the case of a Teacher-Student Perceptron on Gaussian data with SGD dynamics; (ii) in the case of MLPs \rev{or pretrained ResNet backbone, trained on AD CIFAR-10; (iii) in the case of a pretrained ResNet backbone followed by PCA and Perceptron, on a smaller anomaly dataset, BTAD \cite{mishra2021}.}

\section{DISCUSSION}
We analyzed the effect of ADI on learning, through exact analytical calculations, which are compatible with experiments in realistic settings.
In addition to the train and test imbalance, $\rhotrain$ and $\rhotest$, 
in ADI we can identify an intrinsic imbalance, $\rho_0$, over which practitioners have no control.
If data generation is unbiased and no rebalancing of the class distributions is performed and the dataset faithfully represents the deployment distribution, then one has $\rho_0=\rhotrain=\rhotest$. Although our analysis stresses the importance of three kinds of imbalance, we highlight that more sources might arise depending on the situation. For example, it was recently highlighted that, in common  architectures, further phenomena akin to class imbalance can occur at the beginning of learning~\cite{francazi:24}.

Varying $\rhotrain$ corresponds to rebalancing the training distribution. 
Since our results are in the asymptotic data limit, they equally represent the effect of both class reweighting and resampling. Note, however, that these two rebalancing strategies influence SGD differently~\cite{francazi:23}.
Our work shows that the value of $\rhotrain$ which maximizes the overlap between teacher and student is generally not 0.5. 
This is consistent with previous empirical work~\cite{kamalov_partial_2022}, where on different kind of architectures it was shown that re-sampling using some $\rhotrain<0.5$ was consistently optimal over a broad range of tasks and models. The case $\rhotrain>0.5$ was however not explored. A trend was observed: as data is initially more abundant (corresponding to larger $\alpha$ for us), more re-sampling can be done.
Our results show that the picture can in general be more complex.
In fact, this deviation from $\rhotrain^*=0.5$ depends non-linearly on $\rho_0$ and on $\alpha$. While $\alpha$ indicates how much data is available in comparison with the model size, in our linear classifier it also indicates the dimensionality of the input space. Therefore, we cannot disentangle whether this effect is due to model size or to input dimensionality.
We also find that the importance of this deviation is amplified in dynamics with a strong noise (\textit{e.g.}~large learning rate), with small-noise dynamics leading to better solutions than larger-noise ones, with a clear separation between two regimes.\\
This asymmetry is at least in part a consequence of the fact that, in ADI, examples from different classes are intrinsically not equally informative. While this asymmetry was, to our knowledge, not observed in previous work on MGI, we believe that similar deviations from $\rhotrain(R^*)=0.5$ can also be observed, in cases where different classes inform differently on the classification boundary (\textit{e.g.}~a class having smaller variance). In particular, in MG classification, we conjecture that this asymmetry could also be observed in the absence of imbalance. In fact, while in ADI, $\rhotrain(R^*)\neq0.5$ is a feature of the imbalance, in MG classification it can be a feature of the data structure.
\rev{
A consequence is that, under any kind of imbalance, even $\rhotrain$ itself may also be considered as a hyper-parameter.
}

From the point of view of what happens to the training landscape when varying $\rhotrain$, we see that it causes smooth variations in the solutions, with no abrupt changes even when values such as $\rhotrain=\rho_0$ or $0.5$ are crossed. We notice such an absence of phase transitions also when tuning $\rho_0$ and $\rhotest$.\\
Varying $\rhotrain$ and $\rhotest$, and the evaluation metrics, informs us on what each metric reproduces. The balanced accuracy seems the best proxy for the teacher overlap $R$, 
while the quantity that best reflects the bias is the precision $p$.

The choice of $\rhotest$ also has an influence on the \textit{way} data is split between training and test set. While it is common practice to split according to the metadata, the data is sometimes stratified in terms of output value~\cite{zubrod:23}. 
Because minority class examples are more informative than majority class examples, the choice of $\rhotest$ can influence not only the meaning of the optimum, but also the sheer value of the measured metrics. For example, the F1 score, despite being macro-averaged, has higher values when calculated with $\rhotest=0.5$ than with $\rhotest=\rho_0$.


\section{CONCLUSION}

\rev{Our analysis of ADI within an exactly solvable model offers both conceptual insights—such as distinguishing ADI from MGI—and practical implications. In particular we challenge the common knowledge of a balanced training set being optimal, and we show that smaller learning rates are less sensitive to imbalance.}

\subsubsection*{Acknowledgments}
We thank E. Loffredo, B. Loureiro, M. Pastore for insightful conversations.
This work was supported by the Swiss National Foundation, SNF grant \# 196902.
This work is supported by a public grant overseen by the French National Research Agency (ANR) through the program UDOPIA, project funded by the ANR-20-THIA-0013-01. This work is supported by the French government under the France 2030 program (PhOM - Graduate School of Physics) with reference ANR-11-IDEX-0003. 

\bibliography{refs/Related Works,refs/marco}
\clearpage

\appendix
\onecolumn
\aistatstitle{Appendix to ``\ourtitle"}

\section{TEACHER-STUDENT SETUP}\label{app:TS}

\begin{figure}[!ht]
\includegraphics[width=.8
\textwidth]{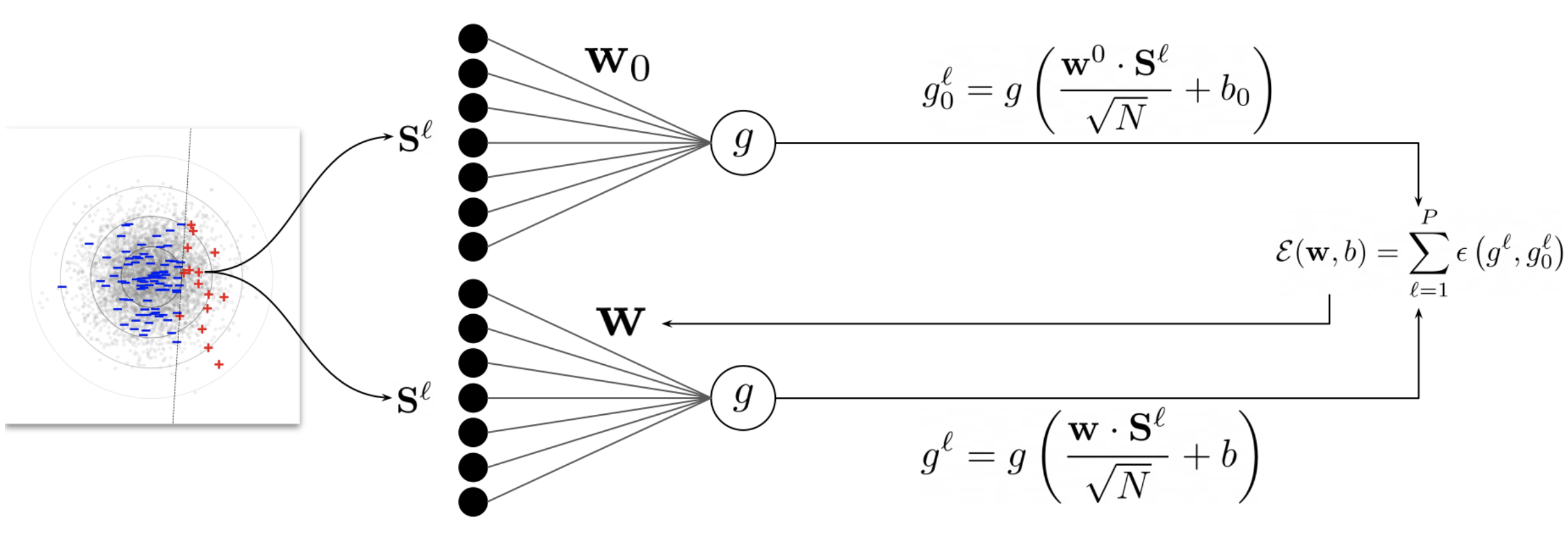}
\caption{Schematic representation of the perceptron model within the teacher-student setup. Labels for the data ${\bf S}^\ell$  are generated by a teacher network with weights ${\bf w}_0$, bias $b_0$ and activation function $g$. The student network (bottom) has the same activation as the teacher (top), and learns its weights ${\bf w}$ and bias $b$ by minimizing a loss $\mathcal{E}$ summed over all data, where $\epsilon$ measures a distance between the label of the data and the student outcome. }
\label{fig:sketch_TS}
\end{figure}

The teacher-student setup has been widely studied in learning theory since the seminal work \cite{gardner1989three} on the perceptron model. 
Figure~\ref{fig:sketch_TS} illustrates this setup, along with the notation used in this work for the teacher and student weights and bias parameters.

\vfill

\section{INFORMATIVE SAMPLES}\label{app:informative}
Here we show that, in the ADI setting, minority class examples are much closer to the teacher's hyperplane (decision boundary) than majority class examples are.
We do this by evaluating the probability of a sample lying on the boundary. Let us consider a sample $\mathbf{S}$ extracted from an $N$-dimensional Gaussian distribution centered at the origin,
\begin{align}
    P(\mathbf{S}) = \frac{1}{(2\pi)^{N/2}}e^{-1/2\sum_{i=1}^N S_i^2}\,.
\end{align}
As we did in the main paper, we study the case $b_0 < 0$ (the positives are the anomalies). We want to compute the probability $P_{boundary}^{+}(b_0)$ that the sample lies between the teacher's hyperplane and the parallel one at a distance $\delta x$, conditioned to $g_0 > 0$. Namely,
\begin{align}
    P_{boundary}^{+}(b_0) \equiv P\left( 0 < \frac{\mathbf{w}_0 \cdot \mathbf{S}}{\sqrt{N}} + b_0 < \delta x \bigg\vert \frac{\mathbf{w}_0 \cdot \mathbf{S}}{\sqrt{N}} + b_0 > 0 \right) &= \frac{P\left( 0 < \frac{\mathbf{w}_0 \cdot \mathbf{S}}{\sqrt{N}} + b_0 < \delta x\right)}{P\left(\frac{\mathbf{w}_0 \cdot \mathbf{S}}{\sqrt{N}} + b_0 > 0\right)}\,.
\end{align}

\begin{figure}[!ht]
    \centering    
    \includegraphics[width=0.5\linewidth]{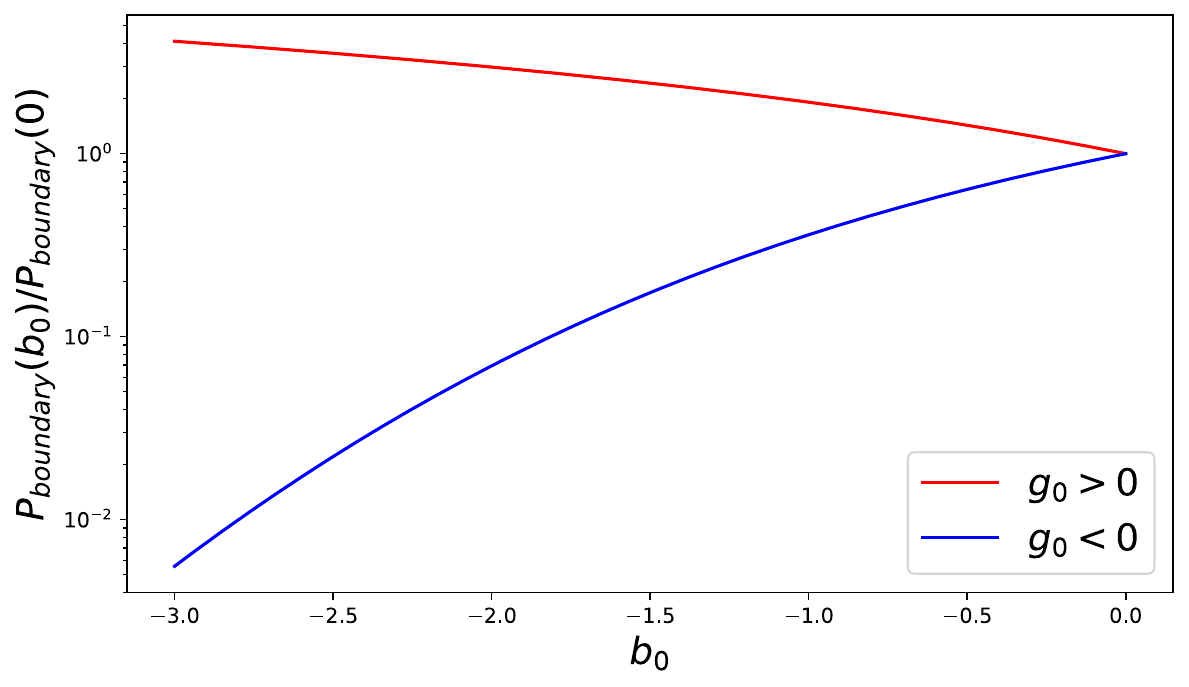}
    \caption{Probability of being close to the boundary (hyperplane), for the minority ($g_0>0$) and for the majority class ($g_0<0$). As a function of $b_0$, we plot the quantities $P_{boundary}^{-}(b_0)$ and $P_{boundary}^{+}(b_0)$, divided by their value at $b_0=0$.}
    \label{fig:app_prob_boundary}
\end{figure}

The denominator is equal to
\begin{align}
    P\left(\frac{\mathbf{w}_0 \cdot \mathbf{S}}{\sqrt{N}} + b_0 > 0\right) &= \int D \mathbf{S} \Theta\left(\frac{\mathbf{w}_0 \cdot \mathbf{S}}{\sqrt{N}} + b_0\right)\\
    &= \int \diff{y} \Theta(y+b_0) \int D\mathbf{S} \delta\left(y  - \frac{\mathbf{w}_0 \cdot \mathbf{S}}{\sqrt{N}}  \right)\,.
\end{align}
By exploiting the Fourier representation of the Dirac $\delta$ distribution one gets
\begin{align}
    P\left(\frac{\mathbf{w}_0 \cdot \mathbf{S}}{\sqrt{N}} + b_0 > 0\right) &=\int \diff{y} \Theta(y+b_0) \int \frac{\diff{\hat{y}}}{2\pi} e^{i\hat{y}y} \int D\mathbf{S} e^{-i\hat{y}\frac{\mathbf{w}_0 \cdot \mathbf{S}}{\sqrt{N}}}\\
    &= \int \diff{y} \Theta(y+b_0) \int \frac{\diff{\hat{y}}}{2\pi} e^{i\hat{y}y-1/2\hat{y}^2}\\
    &= \int \Theta(y+b_0) Dy = \frac{1}{2}\text{Erfc}\left(\frac{-b_0}{\sqrt{2}} \right)\,.
\end{align}
As for the numerator, the computation follows the same lines and one obtains
\begin{align}
    P\left( 0 < \frac{\mathbf{w}_0 \cdot \mathbf{S}}{\sqrt{N}} + b_0 < \delta x\right) = \int_{-b_0}^{-b_0+\delta x}  Dy \xrightarrow{\delta x \to 0} \frac{1}{\sqrt{2\pi}} e^{-b_0^2/2} \delta x\,.
\end{align}
Similarly, one can compute the probability $P_{boundary}^{-}(b_0)$ for samples with $g_0 < 0$, and finally get
\begin{align}
    P_{boundary}^{+}(b_0) &= \frac{\frac{1}{\sqrt{2\pi}} e^{-b_0^2/2} \delta x}{\frac{1}{2}\text{Erfc}\left(\frac{-b_0}{\sqrt{2}} \right)}\,,\\
    P_{boundary}^{-}(b_0) &= \frac{\frac{1}{\sqrt{2\pi}} e^{-b_0^2/2} \delta x}{1 - \frac{1}{2}\text{Erfc}\left(\frac{-b_0}{\sqrt{2}} \right)}\,.
\end{align}
These two quantities are shown in Fig.~\ref{fig:app_prob_boundary} as a function of $b_0 < 0$ (in order to eliminate the dependence on $\delta x$, we may use the corresponding densities; and to set the scale we normalize these quantities by their value at $b_0=0$, which has the property $P_{boundary}^{-}(0)=P_{boundary}^{+}(0)$). It is clear that positive samples have a higher probability to lie on the boundary (hyperplane) as long as the teacher bias is negative, so they represent the minority class in the population distribution.
It is noteworthy that these quantities are actually independent of $N$.

\section{STATISTICAL MECHANICS AND REPLICA CALCULATION}
\label{app:sec:replica_calculations}

\subsection{Statistical mechanics setting}
Statistical mechanics allows to infer macroscopic properties from a large number of interacting agents. In our case, the agents are the model weights, and the properties are quantities such as the performance metrics.

The main quantity we want to calculate is the free energy, $F$, which is a function of the model weights. The minima of $F$ indicate the typical configurations that are assumed by the trained model~\cite{huang:87}.

To calculate the free energy, one usually first calculates the partition function,
\begin{equation}
    Z = \int \diff \mu (\mathbf{w}) \int \diff \mu(b) e ^ {-\beta \mathcal{E}(\mathbf{w},b)}\,.
\end{equation}

From the partition function, one can obtain the free energy through:
\begin{equation}\label{eq:F}
    F = -\frac1\beta \llangle\log Z \rrangle_{\mu_{\text{train}}}\,,
\end{equation}
where $\beta=1/T$ is the inverse temperature.

Eq.~\eqref{eq:F} is formally simple, but it involves calculating an integral of the logarithm of a non-trivial function, which is a hard-to-tract problem.
Replica theory allows us to calculate $F$ by calculating the average of $Z^n$ (where $n$ is a parameter), instead of the average of $\log Z$.

\subsection{Replica Calculation}
The goal of the Replica Calculation is to compute the \textit{Quenched} Free-Energy of the system~\cite{mezard:87},
\begin{equation}
\label{eq_app:free_energy}
    -\beta F = \llangle\log Z \rrangle_{\mu_{\text{train}}} = \llangle \log\int \diff \mu (\mathbf{w}) \int \diff \mu(b) e ^ {-\beta \mathcal{E}(\mathbf{w},b)} \rrangle_{\mu_{\text{train}}}\,,
\end{equation}
where $Z$ is the partition function, and $\llangle \dots \rrangle_{\mu_{\text{train}}} = \int \diff \mu_{\text{train}} (
\{\mathbf{S}^\ell\})$ denotes the average over the distribution of the training data. $\mathcal{E}$ is the training loss as defined in the main text,
\begin{equation}
    \mathcal{E}(\mathbf{w},b) = \sum_{\ell=1}^{P} \epsilon \left(  g \left(  \frac{\mathbf{w}\cdot \mathbf{S}^\ell}{\sqrt{N}} + b  \right), g_0^\ell\right)\,,
\end{equation}
with $\epsilon(x,y) = \frac{1}{2}(x-y)^2$, square loss and $P$ number of training samples. We will also use another shorthand notation for the loss, $\mathcal{E}(\mathbf{w},b) = \sum_{\ell=1}^{N\alpha} \epsilon(\mathbf{w},b;\mathbf{S}^\ell)$, to denote the dependence of term $\ell$ in the sum on the student weights and on the sample $\ell$. The quantity $\diff \mu (\mathbf{w})$ denotes the integration measure over the student weights and it enforces the spherical constraint: 
\begin{equation}
    \diff \mu (\mathbf{w}) = \prod_{i=1}^N \frac{\diff w_i}{\sqrt{2\pi e}} \delta(\mathbf{w}\cdot\mathbf{w}-N)\,.
\end{equation}
The differential $\diff \mu (b) = \diff b$ is the integration measure over the student bias.

We recall the shape of the training data distribution,\footnote{\textbf{Note}: To make the notation lighter, we abbreviate $\rho_{\text{train}}$ as $\rho$ for the whole derivation, since we are now only focusing on the training set.} 
\begin{equation}
    \diff \mu_{\text{train}} (\{\mathbf{S}^\ell\}) = \frac{1}{\mathcal{N}_{\text{bias}}} \binom{N\alpha}{N\alpha\rho} \left(\prod_{\ell=1}^{\alpha N \rho} D\mathbf{S}^\ell \Theta\left(  \frac{\mathbf{w}^0\cdot \mathbf{S}^\ell}{\sqrt{N}} + b_0  \right) \right)\left(\prod_{\ell'=\alpha N \rho + 1}^{\alpha N} D\mathbf{S}^{\ell'} \Theta\left(  -\frac{\mathbf{w}^0\cdot \mathbf{S}^{\ell'}}{\sqrt{N}} - b_0  \right) \right)\,,
\end{equation}
with $\mathcal{N}_{\text{bias}} = \binom{N\alpha}{N\alpha\rho} \left(\frac{1}{2}\text{erfc}\left(\frac{-b_0}{\sqrt{2}}\right) \right)^{\alpha N \rho} \left(1 - \frac{1}{2}\text{erfc}\left(\frac{-b_0}{\sqrt{2}}\right) \right)^{\alpha N (1-\rho)} = \binom{N\alpha}{N\alpha\rho} c_+^{\alpha N \rho} c_-^{\alpha N (1-\rho)}$ normalization constant.

The core of the Replica Approach \cite{mezard_spin_1987} \cite{seung_statistical_1992} lies in exploiting the identity 
\begin{equation}
\label{eq:replica_trick}
    \llangle\log Z \rrangle = \lim_{n\to 0} \frac{1}{n}\log \llangle Z^n\rrangle\,,
\end{equation}
where $n$ is promoted to an integer and $Z^n$ is computed by replicating the partition function $n$ times (\textit{i.e.}~considering $n$ independent copies of the original system). Finally one takes the limit $n\to0$ to recover $\llangle\log Z \rrangle$. This procedure provides exact asympotic results, and was widely used both in machine learning and in several other disordered systems~\cite{charbonneau:23}.

We begin with computing
\begin{align}
    \llangle Z^n \rrangle &= \bigintssss \diff \mu_{\text{train}} (\{\mathbf{S}^\ell\})  \bigintssss \left(\prod_{\sigma=1}^n \diff \mu (b_\sigma) \diff \mu (\mathbf{w}^\sigma) \right) e^{-\beta \sum_\sigma \sum_\ell \epsilon(\mathbf{w}^\sigma,b_\sigma;\mathbf{S}^\ell)}\,.
\end{align}
By expanding the integration measure over the training samples, and collecting respectively positive and negative terms we get, 
\begin{multline}
    \llangle Z^n \rrangle = \bigintssss \prod_{\sigma=1}^n \diff \mu (b_\sigma) \diff \mu (\mathbf{w}^\sigma)  \left[\frac{1}{c_+} \int D\mathbf{S} \Theta\left(  \frac{\mathbf{w}^0\cdot \mathbf{S}}{\sqrt{N}} + b_0  \right) e^{-\beta \sum_\sigma \epsilon(\mathbf{w}^\sigma,b_\sigma;\mathbf{S})} \right]^{N\alpha\rho}\cdot\\
    \cdot\left[\frac{1}{c_-} \int D\mathbf{S} \Theta\left(-\frac{\mathbf{w}^0\cdot \mathbf{S}}{\sqrt{N}} - b_0  \right) e^{-\beta \sum_\sigma \epsilon(\mathbf{w}^\sigma,b_\sigma;\mathbf{S})} \right]^{N\alpha(1-\rho)} \,.
    \label{eq:need_of_b}
\end{multline}
Now we define
\begin{align}
\label{eq:energetic_terms_separate}
    \mathcal{G}_r^\pm (\{ \mathbf{w}^\sigma,b_\sigma\}) &= - \log \frac{1}{c_\pm} \int D \mathbf{S} \Theta\left(\pm\frac{\mathbf{w}^0\cdot \mathbf{S}}{\sqrt{N}} \pm b_0  \right) e^{-\beta \sum_\sigma \epsilon(\mathbf{w}^\sigma,b_\sigma;\mathbf{S})}\,,
\end{align}
where $\{ \mathbf{w}^\sigma,b_\sigma\}$ denotes the dependence of $\mathcal{G}_r^\pm$ on the whole set of $n$ replicated students. Then one can rewrite the replicated partition function as: 
\begin{align}
\label{eq:energetic_terms}
    \llangle Z^n \rrangle &= \bigintssss \prod_{\sigma=1}^n \diff \mu (b_\sigma) \diff \mu (\mathbf{w}^\sigma) e^{-N\alpha\rho \mathcal{G}_r^+ (\{ \mathbf{w}^\sigma,b_\sigma\}) - N\alpha(1-\rho)\mathcal{G}_r^- (\{ \mathbf{w}^\sigma,b_\sigma\})}\,.
\end{align}
We introduce the auxiliary variables $x_\sigma$ and $y$ in order to remove $\mathbf{S}$ from the argument of $g$ function in the expression of $\mathcal{G}_r^\pm$: 
\begin{align}
    e^{-\mathcal{G}_r^+} &= \frac{1}{c_+}\int \prod_{\sigma=1}^n \diff x_\sigma \int \diff y \; \Theta(y+b_0)e^{-\frac{\beta}{2}\sum_\sigma\left[ g(x_\sigma+b_\sigma) - g(y+b_0)   \right]^2 } \int D \mathbf{S} \prod_{\sigma=1}^n \delta\left(x_\sigma - \frac{\mathbf{w}^\sigma\cdot \mathbf{S}}{\sqrt{N}}\right) \delta\left(y - \frac{\mathbf{w}^0\cdot \mathbf{S}}{\sqrt{N}}\right)\\
    &=\frac{1}{c_+}\bigintssss \prod_{\sigma=1}^n \frac{\diff x_\sigma \diff \hat{x}_\sigma}{2\pi} \bigintssss \frac{\diff y \diff \hat{y}}{2\pi} \Theta(y+b_0) e^{\left( -\frac{\beta}{2}\sum_\sigma\left[ g(x_\sigma+b_\sigma) - g(y+b_0)   \right]^2 + i \sum_\sigma x_\sigma \hat{x}_\sigma + i y \hat{y} \right)} \int D \mathbf{S} e^{-i \left(  \sum_\sigma \mathbf{w}^\sigma \hat{x}_\sigma + \mathbf{w}^0 \hat{y}  \right)\cdot \frac{\mathbf{S}}{\sqrt{N}}},\label{eq:last}
\end{align}
where we exploited the Fourier representation of the $\delta$-function. 
The last integral in Eq.~\eqref{eq:last} is Gaussian, and it yields: $e^{-\frac{1}{2N}(\sum_\sigma \mathbf{w}^\sigma \hat{x}_\sigma + \mathbf{w}^0\hat{y})^2}$. The function $\mathcal{G}_r^+$  depends on the vectors ${\bf w}^\sigma, {\bf w}^0$ only through the overlaps 
\begin{align}
    Q_{\sigma \sigma'} = \frac{\mathbf{w}^\sigma\cdot \mathbf{w}^{\sigma'}}{N}, \;\;\; R_\sigma = \frac{\mathbf{w}^\sigma\cdot\mathbf{w}^0}{N},
\end{align}
which are emergent order parameters of the theory. In terms of these functions, $\mathcal{G}_r^+$ can be written as
\begin{align}
    e^{-\mathcal{G}_r^+} &=\frac{1}{c_+} \bigintssss \prod_{\sigma=1}^n \frac{\diff x_\sigma \diff \hat{x}_\sigma}{2\pi} \bigintssss \frac{\diff y \diff \hat{y}}{2\pi} \Theta(y+b_0) e^{-\frac{\beta}{2}\sum_\sigma\left[ g(x_\sigma+b_\sigma) - g(y+b_0)   \right]^2 + i \sum_\sigma x_\sigma \hat{x}_\sigma + i y \hat{y}} e^{-\frac{1}{2}\sum_{\sigma,\sigma'} \hat{x}_\sigma\hat{x}_{\sigma'} Q_{\sigma\sigma'} - \hat{y}\sum_\sigma \hat{x}_\sigma R_\sigma - \frac{1}{2}\hat{y}^2},
\end{align}
and similarly
\begin{align}
    e^{-\mathcal{G}_r^-} &= \frac{1}{c_-}\bigintssss \prod_{\sigma=1}^n \frac{\diff x_\sigma \diff \hat{x}_\sigma}{2\pi} \bigintssss \frac{\diff y \diff \hat{y}}{2\pi} \Theta(-y-b_0) e^{ -\frac{\beta}{2}\sum_\sigma\left[ g(x_\sigma+b_\sigma) - g(y+b_0)   \right]^2 + i \sum_\sigma x_\sigma \hat{x}_\sigma + i y \hat{y} } e^{-\frac{1}{2}\sum_{\sigma,\sigma'} \hat{x}_\sigma\hat{x}_{\sigma'} Q_{\sigma\sigma'} - \hat{y}\sum_\sigma \hat{x}_\sigma R_\sigma - \frac{1}{2}\hat{y}^2}.
\end{align}
The replicated partition function can  thus be written as an integral over the order parameters:
\begin{align}
    \llangle Z^n \rrangle &= \bigintssss \prod_{\sigma=1}^n \diff \mu (b_\sigma) \diff \mu (\mathbf{w}^\sigma) e^{-N\alpha\rho \mathcal{G}_r^+(\{ \mathbf{w}^\sigma, b_\sigma \}) - N\alpha(1-\rho)\mathcal{G}_r^- (\{ \mathbf{w}^\sigma, b_\sigma \})}\times\nonumber\\
    &\hspace{50pt}\times\bigintssss \prod_{\sigma > \sigma'} \diff Q_{\sigma\sigma'} \bigintssss \prod_{\sigma} \diff R{\sigma} \prod_{\sigma > \sigma'} \delta(\mathbf{w}^{\sigma}\cdot\mathbf{w}^{\sigma'} - N Q_{\sigma \sigma'} ) \prod_{\sigma} \delta(\mathbf{w}^{\sigma}\cdot\mathbf{w}^{0} - N R_{\sigma})\\ 
     &= \bigintssss \prod_{\sigma} \diff b_\sigma \bigintssss \prod_{\sigma > \sigma'} \frac{\diff Q_{\sigma\sigma'} \diff \hat{Q}_{\sigma\sigma'}}{2\pi i} \bigintssss \prod_{\sigma} \frac{\diff R_{\sigma} \diff \hat{R}_{\sigma}}{2\pi i}e^{-N\alpha\rho \mathcal{G}_r^+(\{Q_{\sigma \sigma'},R_\sigma, b_\sigma\}) - N\alpha(1-\rho)\mathcal{G}_r^- (\{Q_{\sigma \sigma'},R_\sigma, b_\sigma\})}\times\nonumber\\
     &\hspace{50pt}\times e^{N\left( -\sum_{\sigma>\sigma'}Q_{\sigma\sigma'}\hat{Q}_{\sigma\sigma'} - \sum_{\sigma} \hat{R}_\sigma R_\sigma   \right)}\bigintssss\prod_{\sigma=1}^n \diff \mu (\mathbf{w}^\sigma)  e^{\sum_{\sigma>\sigma'}\hat{Q}_{\sigma\sigma'} \mathbf{w}^\sigma \mathbf{w}^{\sigma'} + \sum_{\sigma} \hat{R}_\sigma \mathbf{w}^\sigma \mathbf{w}^{0}}   \,,
\end{align}
with $\hat{Q}_{\sigma\sigma'}$ and $\hat{R}_\sigma$ conjugates of the overlaps.
Then, the replicated partition function can be rewritten as 
\begin{align}
\label{eq:Zn_energetic+volume}
    \llangle Z^n \rrangle = \bigintssss \prod_{\sigma} \diff b_\sigma \bigintssss \prod_{\sigma > \sigma'} \frac{\diff Q_{\sigma \sigma'} \diff \hat{Q}_{\sigma\sigma'}}{2\pi i} \bigintssss \prod_{\sigma} \frac{\diff R_{\sigma} \diff \hat{R}_{\sigma}}{2\pi i}e^{-N \mathcal{A}_r(\{Q_{\sigma\sigma'},\hat{Q}_{\sigma\sigma'},R_\sigma,\hat{R}_\sigma\})}\,,
\end{align}
where we defined
\begin{align}\label{eq:Action}
   \mathcal{A}_r(\{Q_{\sigma\sigma'},\hat{Q}_{\sigma\sigma'},R_\sigma,\hat{R}_\sigma\})=\alpha\rho \mathcal{G}_r^+(\{Q_{\sigma\sigma'},R_\sigma , b_\sigma\})+\alpha(1-\rho)\mathcal{G}_r^- (\{Q_{\sigma\sigma'},R_\sigma, b_\sigma \})- \mathcal{G}_0(\{Q_{\sigma\sigma'},\hat{Q}_{\sigma\sigma'},R_\sigma,\hat{R}_\sigma\})\,,
\end{align}
and 
\begin{align}
\label{eq:G0_beforeRS}
    \mathcal{G}_0(\{Q_{\sigma\sigma'},\hat{Q}_{\sigma\sigma'},R_\sigma,\hat{R}_\sigma\}) =  -\sum_{\sigma>\sigma'}Q_{\sigma \sigma'}\hat{Q}_{\sigma\sigma'} - \sum_{\sigma} \hat{R}_\sigma R_\sigma + \frac{1}{N} \log \bigintssss\prod_{\sigma=1}^n \diff \mu (\mathbf{w}^\sigma)  e^{\sum_{\sigma>\sigma'}\hat{Q}_{\sigma\sigma'} \mathbf{w}^\sigma \mathbf{w}^{\sigma'} + \sum_{\sigma} \hat{R}_\sigma \mathbf{w}^\sigma \mathbf{w}^{0}}.
\end{align}

\paragraph{Replica Symmetric Ansatz}
According to Eq.~\eqref{eq:replica_trick}, the Free Energy density in the thermodynamic limit ($N\to\infty$) reads:
\begin{equation}
    -\beta \mathcal{F} =  \lim_{N\to\infty} \lim_{n\to 0} \frac{1}{nN} \log \llangle Z^n \rrangle\,.
\end{equation}
In order to evaluate the integral in Eq.~\eqref{eq:Zn_energetic+volume} by saddle-point we switch the order of the two limits as prescribed within replica theory, getting,
\begin{equation}
\label{eq:saddle_point}
    -\beta \mathcal{F} = \lim_{n\to 0} \frac{1}{n} \min\limits_{\substack{Q_{\sigma \sigma'},R_\sigma,\\\hat{Q}_{\sigma \sigma'},\hat{R}_\sigma, b_\sigma}}\left\{ \mathcal{G}_0(\{Q_{\sigma \sigma'},\hat{Q}_{\sigma \sigma'},R_\sigma,\hat{R}_\sigma\})  -\alpha\rho \mathcal{G}_r^+(\{Q_{\sigma \sigma'},R_\sigma,b_\sigma\}) - \alpha(1-\rho)\mathcal{G}_r^- (\{Q_{\sigma \sigma'},R_\sigma,b_\sigma\}) \right\}.
\end{equation}
To carry on the computation one has to find a parametrization of the order parameters and express Eq.~\eqref{eq:saddle_point} as a function of the elements of these multi-dimensional arrays and the number of replicas $n$. We adopt the Replica Symmetric (RS) ansatz~\cite{mezard:87}, where one assumes that the replicated students are symmetric, \textit{i.e.}~they all have the same overlap with the teacher and among them: 
\begin{align}
Q_{\sigma\sigma'} =& \delta_{\sigma\sigma'} + (1-\delta_{\sigma\sigma'})q\,,\\
\hat{Q}_{\sigma\sigma'} =& \delta_{\sigma\sigma'} + (1-\delta_{\sigma\sigma'})\hat{q}\,,\\
R_\sigma =& R\,,\\
\hat{R}_\sigma =& \hat{R}\,,\\
b_\sigma =& b\,.
\end{align}
We stress that in our computation the student bias $b$ is treated as an order parameter and its value is fixed by saddle point as for the other parameters. We now define: 
\begin{align}
    G_r^\pm = \lim_{n\to0} \frac{\mathcal{G}_r^\pm}{n}, \;\;\;\;\; G_0 = \lim_{n\to0}\frac{\mathcal{G}_0}{n}\,.
\end{align}
At this point, the optimization problem to solve in order to find the equilibrium values of the order parameters becomes the following: 
\begin{align}
    -\beta \mathcal{F} = \min\limits_{\substack{q,R,\\\hat{q},\hat{R},b}}\left\{ G_0(q,\hat{q},R,\hat{R}) -\alpha\rho G_r^+(q,R,b) -\alpha(1-\rho) G_r^-(q,R,b) \right\}\,.
\end{align}
We will refer to $G_r^\pm$ as the energetic terms as they represent the energetic contribution to the free-energy of positive and negative class samples. $G_0$ is the entropic or volume term and quantifies the number of student configurations that correspond to a given choice of the order parameters.

In the following, we show the detailed computations for $G_r^+$, as the derivation  for $G_r^-$ follows the same lines. The plan is to substitute the RS ansatz in the expression of $\mathcal{G}_r^+$ and integrate over the conjugate variables $\hat{y},\hat{x}_\sigma$. The integral in $\hat{y}$ is a Gaussian integral and yields: 
\begin{align}
    e^{-\mathcal{G}_r^+} &= \frac{1}{c_+} \bigintssss \prod_{\sigma=1}^n \frac{\diff x_\sigma \diff \hat{x}_\sigma}{2\pi} \bigintssss \frac{\diff y}{\sqrt{2\pi}} e^{-\frac{y^2}{2}} \Theta(y+b_0)\nonumber\\
    &\hspace{50pt} e^{ -\frac{\beta}{2}\sum_\sigma\left[ g(x_\sigma+b) - g(y+b_0)   \right]^2 } e^{-\frac{1}{2}(1-q)\sum_\sigma \hat{x}_\sigma^2 +\frac{1}{2}(R^2 - q)\sum_{\sigma,\sigma'} \hat{x}_\sigma\hat{x}_{\sigma'} + i\sum_\sigma \hat{x}_\sigma(x_\sigma - yR)}\,.
\end{align}

In order to integrate out the $\hat{x}_\sigma$ we need to decouple the term $\hat{x}_{\sigma}\hat{x}_{\sigma'}$ through Hubbard-Stratonovich transform,
\begin{equation}
    e^{-\frac{1}{2}(q-R^2)\sum_{\sigma,\sigma'} \hat{x}_\sigma\hat{x}_{\sigma'}} = \int D t e^{(i\sqrt{q-R^2}\sum_\sigma\hat{x}_\sigma)t }\,,
\end{equation}
where we recall that $D t = \frac{\diff t}{\sqrt{2\pi}}e^{-t^2/2}$.

Integrals over $\hat{x}_\sigma$ are now Gaussian and yield:
\begin{align}
    e^{-\mathcal{G}_r^+} = \frac{1}{c_+} \bigintssss D y \bigintssss D t \Theta(y+b_0) \prod_{\sigma=1}^n \bigintssss \frac{\diff x_\sigma}{\sqrt{2\pi(1-q)}} e^{-\frac{(x_\sigma - yR + t\sqrt{q-R^2})^2}{2(1-q)}}  e^{ -\frac{\beta}{2}\left[ g(x_\sigma+b) - g(y+b_0)   \right]^2}\,.
\end{align}

Performing the shift and re-scaling $x_\sigma \to x_\sigma\sqrt{1-q} + yR - t\sqrt{q-R^2}$ one gets:
\begin{align}
    e^{-\mathcal{G}_r^+} = \frac{1}{c_+} \bigintssss D y \Theta(y+b_0) \bigintssss D t  \left[ \bigintssss Dx  e^{ -\frac{\beta}{2}\left[ g(x\sqrt{1-q} + yR - t\sqrt{q-R^2}+b) - g(y+b_0)   \right]^2} \right]^n\,.
\end{align}

Now we can compute the $n\to0$ limit. We call $A = \bigintssss Dx  e^{ -\frac{\beta}{2}\left[ g(x\sqrt{1-q} + yR - t\sqrt{q-R^2}+b) - g(y+b_0)   \right]^2}$, and exploit the identity $A^n \sim 1 + n \log A$, which is valid in the $n\to0$ limit:
\begin{align}
    G_r^+ = \lim_{n\to 0}- \frac{1}{n} \log \left(1 + n \frac{1}{c_+} \bigintssss D y \Theta(y+b_0) \bigintssss D t   \log A    \right)\,.
\end{align}

Since $n$ is small, we can expand the first logarithm around 1:
\begin{align}
    G_r^+ = - \frac{1}{c_+} \bigintssss D y \Theta(y+b_0) \bigintssss D t   \log A\,.
\end{align}

We now recall our choice of the activation function $g(x) = \text{sign}(x)$. The square-loss per sample in this case reads $\epsilon(\mathbf{w};\mathbf{S}) = 2 \Theta(-(\mathbf{w}\cdot\mathbf{S}+b)(\mathbf{w}^0\cdot\mathbf{S}+b_0))$. We re-define it without the the factor $2$ in order to count the number of mis-classified samples. $G_r^+$ becomes
\begin{align}
    G_r^+ &= - \frac{1}{c_+} \bigintssss_{-b_0}^\infty D y  \bigintssss_{-\infty}^\infty D t   \log \left( \bigintssss_{-\infty}^\infty Dx  e^{ -\beta\Theta\left( -(x\sqrt{1-q} + yR - t\sqrt{q-R^2}+b)(y+b_0)\right)}  \right)\,.
\end{align}

We define $u \equiv \frac{t\sqrt{q-R^2}-yR-b}{\sqrt{1-q}}$ and $H(x) \equiv \int_x^\infty D t = \frac{1}{2}\text{erfc}\left(\frac{x}{\sqrt{2}}\right)$. The final form for $G_r^\pm$ reads
\begin{align}
\label{eq:Gr_plus_Gr_minus}
    G_r^+ &=- \frac{1}{c_+} \bigintssss_{-b_0}^\infty D y \bigintssss_{-\infty}^\infty D t  \log \left( e^{-\beta} + (1-e^{-\beta}) H(u) \right)\,,\\
    G_r^- &= - \frac{1}{c_-} \bigintssss_{-\infty}^{-b_0} D y \bigintssss_{-\infty}^\infty D t  \log \left((e^{-\beta}-1) H(u)+1 \right)\,.
\end{align}

We now show the detailed computation for the entropic term $G_0$. Starting from Eq.~\eqref{eq:G0_beforeRS}, and substituting the RS ansatz one gets
\begin{align}
    \mathcal{G}_0 =  -\frac{1}{2}n(n-1)q\hat{q} - n \hat{R} R + \frac{1}{N} \log \bigintssss\prod_{\sigma=1}^n \diff \mu (\mathbf{w}^\sigma)  e^{\hat{q}\sum_{\sigma>\sigma'}\mathbf{w}^\sigma \mathbf{w}^{\sigma'} + \hat{R}\sum_{\sigma} \mathbf{w}^\sigma \mathbf{w}^{0}}\,.
\end{align}

We decouple $\mathbf{w}^\sigma \mathbf{w}^{\sigma'}$ through Hubbard-Stratonovich:
\begin{align}
e^{\hat{q}\sum_{\sigma>\sigma'}\mathbf{w}^\sigma \mathbf{w}^{\sigma'}} = e^{\frac{1}{2}\hat{q}\sum_{\sigma,\sigma'}\mathbf{w}^\sigma \mathbf{w}^{\sigma'} - \frac{1}{2}\hat{q}\sum_{\sigma}\mathbf{w}^\sigma \mathbf{w}^{\sigma}} = e^{- \frac{1}{2}\hat{q}\sum_{\sigma}\mathbf{w}^\sigma \mathbf{w}^{\sigma}}\int D \mathbf{z} e^{\sqrt{\hat{q}}\sum_\sigma \mathbf{w}^\sigma \mathbf{z}}\,.
\end{align}

This allows us to rewrite $\mathcal{G}_0$ as
\begin{align}
    \mathcal{G}_0 &= -\frac{1}{2}n(n-1)q\hat{q} - n \hat{R} R + \frac{1}{N}\log \int D \mathbf{z} \left(\int \diff \mu (\mathbf{w}) e^{\mathbf{w}(\hat{R}\mathbf{w}^0+ \sqrt{\hat{q}}\mathbf{z}- \frac{1}{2}\hat{q}\mathbf{w})}\right)^n\,.
\end{align}

Now we can take the $n\to0$ limit: 
\begin{align}
    G_0 = \lim_{n\to0} \frac{\mathcal{G}_0}{n} = - \hat{R}R + \frac{1}{2}q\hat{q} + \frac{1}{N} \int D \mathbf{z} \log \int \diff \mu (\mathbf{w}) e^{\mathbf{w}(\hat{R}\mathbf{w}^0+ \sqrt{\hat{q}}\mathbf{z}- \frac{1}{2}\hat{q}\mathbf{w})}.\label{eq:last_int}
\end{align}

The last integral on Eq.~\eqref{eq:last_int} is equal to: 
\begin{align}
    \int D \mathbf{z} \log \int \frac{\diff \lambda}{4\pi i}e^{ \frac{N \lambda}{2}} e^{-\frac{N}{2}\log[ e (\lambda+\hat{q}])} e^ {\frac{N}{2(\lambda+\hat{q})}(\hat{R}^2 + \hat{q}\frac{\sum_i z_i^2}{N} + 2 \sqrt{\hat q} \hat R \frac{\sum_i \omega^0_i z_i}{N} )}.
\end{align}
Computing the integral over $\lambda$ with a saddle point approximation we reduce the double integral to
\begin{align}
   N \left[ \frac{ \lambda}{2} -\frac{1}{2}\log[ e (\lambda+\hat{q})] +\frac{\hat{R}^2}{2(\lambda+\hat{q})} +\frac{1}{2(\lambda+\hat{q})}\int D \mathbf{z}  \left(\hat{q}\frac{\sum_i z_i^2}{N} + 2 \sqrt{\hat q} \hat R \frac{\sum_i \omega^0_i z_i}{N} \right)\right], 
\end{align}
where now $\lambda$ denotes the saddle point value. The Gaussian integral over ${\bf z}$ is easily computed, and one gets as a result: 
\begin{align}
    G_0 &=
     - \hat{R}R + \frac{1}{2}q\hat{q} + \frac{\lambda}{2} - \frac{1}{2}\log(\lambda+\hat{q}) + \frac{1}{2} \frac{\hat{R}^2 + \hat{q}}{(\lambda + \hat{q})}  - \frac{1}{2}.
\end{align}

\paragraph{Saddle Point Equations}
We look for a stationary point of the variational Free-Energy to fix the value of the order parameters at equilibrium. We then set to $0$ the derivatives of the variational Free-Energy with respect to the order parameters ($q,\hat{q},R,\hat{R},b$) and the additional Lagrange multiplier $\lambda$ that we introduced to enforce the spherical constraint for the students' weights. By doing so, we obtain the following system of coupled equations:
\begin{align}
    R &= \hat{R}(1-q)\,,\label{eq:saddle1}\\
    q &= (\hat{q}+\hat{R}^2)(1-q)^2\,,\label{eq:saddle2}\\
    \hat{R} &= \alpha\frac{ e^{-\frac{b_0^2 q}{2(q-R^2)}}}{2\pi\sqrt{1-q}} \left\{ \frac{\rho}{c_+} \bigintssss_{-\infty}^{\infty} D t \frac{e^{-v^2/2+\frac{b_0 R}{\sqrt{q-R^2}}t}}{(e^\beta - 1)^{-1} + H(v)} - \frac{1-\rho}{c_-}\bigintssss_{-\infty}^{\infty} D t \frac{e^{-v^2/2+\frac{b_0 R}{\sqrt{q-R^2}}t}}{(e^{-\beta} - 1)^{-1} + H(v)}\right\}\,,\label{eq:saddle3}\\
    \hat{q} &= \frac{\alpha}{2\pi(1-q)}  \left\{  \frac{\rho}{c_+} \bigintssss_{-b_0}^{\infty} D y \bigintssss_{-\infty}^{\infty} D t \frac{e^{-u^2}}{[(e^\beta - 1)^{-1} + H(u)]^2} + \frac{1-\rho}{c_-} \bigintssss_{-\infty}^{-b_0} D y \bigintssss_{-\infty}^{\infty} D t \frac{e^{-u^2}}{[(e^{-\beta} - 1)^{-1} + H(u)]^2}  \right\}\,,\label{eq:saddle4}\\
    0 &=\frac{\rho}{c_+} \bigintssss_{-b_0}^{\infty} D y \bigintssss_{-\infty}^{\infty} D t  \frac{e^{-u^2/2}}{(e^{\beta}-1)^{-1}+H(u)} + \frac{1-\rho}{c_-} \bigintssss_{-\infty}^{-b_0} D y \bigintssss_{-\infty}^{\infty} D t  \frac{e^{-u^2/2}}{(e^{-\beta}-1)^{-1}+H(u)}\,.\label{eq:saddle5}
\end{align}
In the following and in the main manuscript, when we talk of self-consistently solving the saddle-point equations, we refer to solving Eqs.~(\ref{eq:saddle1}--\ref{eq:saddle5}).

\section{TRAIN AND GENERALIZATION METRICS}
\label{app:sec:train_and_generalization_metrics}

\paragraph{Analytical expression for the Metrics}
Train and Generalization metrics can be expressed in terms of order parameters evaluated at equilibrium (solutions of the saddle-point equations). Here we derive their expressions. 

The average \textbf{train error} per sample can be evaluated as: 
\begin{align}
    \epsilon_t = \frac{1}{P} \llangle \mathbb{E}_T[\mathcal{E}(\mathbf{w},b)]\rrangle_{\mu_\text{train}}\,,
\end{align}
where $\mathbb{E}_T[\dots]$ denotes the average over the realizations of the thermal noise. Explicitly, one has
\begin{align}
    \epsilon_t = \frac{1}{N\alpha} \llangle \frac{1}{Z} \int \diff \mu (\mathbf{w}) \int \diff \mu(b) \mathcal{E}(\mathbf{w},b) e ^ {-\beta \mathcal{E}(\mathbf{w},b)} \rrangle_{\mu_\text{train}} = - \frac{1}{N\alpha} \llangle \frac{\partial}{\partial\beta} \log Z \rrangle = \frac{1}{N\alpha} \frac{\partial (\beta F)}{\partial\beta}\,,
\end{align}

\begin{align}
    \epsilon_t = -\frac{1}{\alpha} \frac{\partial}{\partial\beta} \left\{ G_0(q,\hat{q},R,\hat{R}) -\alpha\rho_{\text{train}} G_r^+(q,R,b) -\alpha(1-\rho_{\text{train}}) G_r^-(q,R,b) \right\}\,,
\end{align}

evaluated at the saddle point. The volume term $G_0$ does not depend on the temperature. We get

\begin{align}
    \epsilon_t &= \rho_{\text{train}} \frac{\partial G_r^+(q,R,b)}{\partial\beta} + (1-\rho_{\text{train}}) \frac{\partial G_r^-(q,R,b)}{\partial\beta}\\
    &= \frac{\rho_{\text{train}}}{c_+} \bigintssss_{-b_0}^\infty D y \bigintssss_{-\infty}^\infty D t \frac{1-H(u)}{1 + (e^\beta - 1) H(u)} + \frac{1-\rho_{\text{train}}}{c_-}  \bigintssss_{-\infty}^{-b_0} D y \bigintssss_{-\infty}^\infty D t \frac{H(u)}{e^\beta + (1-e^\beta) H(u)}\,.
\end{align}

To compute the generalization metrics we introduce the test-set distribution
\begin{align}
    \diff \mu_{\text{test}} (\mathbf{S}) = \frac{\rho_{\text{test}}}{c_+}&\Theta\left(  \frac{\mathbf{w}^0\cdot \mathbf{S}}{\sqrt{N}} + b_0  \right) D\mathbf{S} +\frac{1-\rho_{\text{test}}}{c_-} \Theta\left(  -\frac{\mathbf{w}^0\cdot \mathbf{S}}{\sqrt{N}} - b_0  \right) D\mathbf{S}\,,
\end{align}
where $\rho_{\text{test}}$ determines the probability of having a positive sample in the test-set. The idea behind the computation of generalization metrics is to add one sample that was not observed during training and evaluate the performance of the trained student on it. In practice, we can define the \textbf{generalization error} as
\begin{align}
    \epsilon_g = \llangle  \llangle \mathbb{E}_T[\epsilon(\mathbf{w},b;\mathbf{S})] \rrangle_{\mu_{\text{train}}}\rrangle_{\mu_{\text{test}}}\,.
\end{align}
The first average is on the train-set and it yields the saddle-point equations shown in the previous section. The second average, on the test-set is needed to evaluate the trained student on the new, unseen sample. Explicitly, we get
\begin{align}
    \epsilon_g &= \llangle  \llangle \frac{1}{Z} \int \diff \mu (\mathbf{w}) \int \diff \mu(b) \epsilon (\mathbf{w},b;\mathbf{S}) e ^ {-\beta \mathcal{E}(\mathbf{w},b)} \rrangle_{\mu_{\text{train}}}\rrangle_{\mu_{\text{test}}}\\
    &= \lim_{n\to 0} \llangle \llangle Z^{n-1} \int \diff \mu (\mathbf{w}) \int \diff \mu(b) \epsilon (\mathbf{w},b;\mathbf{S}) e^{-\beta \sum_{\ell = 1}^{N\alpha}\epsilon (\mathbf{w},b;\mathbf{S}^\ell)} \rrangle_{\mu_{\text{train}}}\rrangle_{\mu_{\text{test}}}\\
    &= \lim_{n\to 0} \bigintssss \diff \mu_{\text{test}}(\mathbf{S}) \bigintssss \diff \mu_{\text{bias}} (\{\mathbf{S}^\ell\}) \bigintssss\prod_{\sigma=1}^n \diff\mu(b_\sigma) \diff\mu(\mathbf{w}^\sigma) \epsilon (\mathbf{w}^1,b_1;\mathbf{S}) e^{-\beta \sum_\ell \sum_\sigma \epsilon (\mathbf{w}^\sigma,b_\sigma;\mathbf{S}^\ell)}\\
    &= \lim_{n\to 0} \bigintssss \diff\mu_{\text{test}}(\mathbf{S}) \bigintssss\prod_{\sigma=1}^n \diff\mu(b_\sigma) \diff\mu(\mathbf{w}^\sigma) \epsilon (\mathbf{w}^1,b_1;\mathbf{S}) e^{-N\alpha\rho \mathcal{G}_r^+ (\{ \mathbf{w}^\sigma,b_\sigma\}) - N\alpha(1-\rho)\mathcal{G}_r^- (\{ \mathbf{w}^\sigma,b_\sigma \})}\,.
\end{align}

One can evaluate:
\begin{align}
    \int \diff\mu_{\text{test}}(\mathbf{S}) \epsilon(\mathbf{w}^1,b_1;\mathbf{S}) = \frac{\rho_{\text{test}}}{c_+} \mathrm{I}_+ + \frac{1-\rho_{\text{test}}}{c_-} \mathrm{I}_-\,,
\end{align}
where we have defined:
\begin{align}
    \mathrm{I}_\pm &= \int D \mathbf{S} \Theta\left(  \pm\frac{\mathbf{w}^0\cdot \mathbf{S}}{\sqrt{N}} \pm b_0  \right) \epsilon(\mathbf{w}^1,b_1;\mathbf{S})\,.
\end{align}

In the following we show the computation for $\mathrm{I}_+$, the one for $\mathrm{I}_-$ follows the same lines:
\begin{align}
    \mathrm{I}_+ &= \int D \mathbf{S} \Theta\left(  \frac{\mathbf{w}^0\cdot \mathbf{S}}{\sqrt{N}} + b_0  \right) \epsilon(\mathbf{w}^1,b_1;\mathbf{S})\\
    &= \int \frac{\diff x \diff\hat{x}}{2\pi} \int \frac{\diff y \diff\hat{y}}{2\pi} e^{i x \hat{x}+i y \hat{y}}\frac{1}{2}[g(x+b_1)-g(y+b_0)]^2 \Theta (y + b) \int D \mathbf{S} e^{-\frac{i}{\sqrt{N}}(\mathbf{w}^1\hat{x}+\mathbf{w}^0 \hat{y})\cdot\mathbf{S}}\,.
\end{align}
Integrating over $\mathbf{S},\hat{x},\hat{y}$ one gets
\begin{align}
    \mathrm{I}_+ &= \int D x \int_{-b_0}^{\infty} D y \frac{1}{2}[g(x\sqrt{1-R_1^2}+yR_1+b_1)-g(y+b_0)]^2 =\\
    &= \int D x \int_{-b_0}^{\infty} D y \Theta(-(x\sqrt{1-R_1^2}+yR_1+b_1)(y+b_0)) =\\
    &= \int_{-b_0}^{\infty} D y \int_{-\infty}^{u'} D x = \int_{-b_0}^{\infty} D y (1-H(u'))\,,
\end{align}
with $u' = \frac{-yR_1 -b_1}{\sqrt{1-R_1^2}}$.

Computing also $\mathrm{I}_-$ we get: 
\begin{align}
    \epsilon(R_1,b_1) &= \frac{\rho_{\text{test}}}{\frac{1}{2}\text{erfc}\left(\frac{-b_0}{\sqrt{2}}\right)} \int_{-b_0}^{\infty} D y \int_{-\infty}^{u'} D x + \frac{1-\rho_{\text{test}}}{1-\frac{1}{2}\text{erfc}\left(\frac{-b_0}{\sqrt{2}}\right)} \int_{-\infty}^{-b_0} D y \int_{u'}^{\infty} D x \,.
\end{align}

Thus we can rewrite the generalization error: 
\begin{align}
    \epsilon_g = \lim_{n\to 0} \bigintssss \prod_{\sigma} \diff b_\sigma \bigintssss \prod_{\sigma > \sigma'} &\frac{\diff Q_{\sigma,\sigma'} \diff \hat{Q}_{\sigma,\sigma'}}{2\pi i} \bigintssss \prod_{\sigma} \frac{\diff R_{\sigma} \diff \hat{R}_{\sigma}}{2\pi i} \times\\
    &\times\epsilon(R_1,b_1) e^{-N\alpha\rho \mathcal{G}_r^+(\{Q_{\sigma,\sigma'},R_\sigma,b_\sigma\}) - N\alpha(1-\rho)\mathcal{G}_r^- (\{Q_{\sigma,\sigma'},R_\sigma,b_\sigma\}) + N\mathcal{G}_0(\{Q_{\sigma,\sigma'},\hat{Q}_{\sigma,\sigma'},R_\sigma,\hat{R}_\sigma\})}\,.
\end{align}

Following the same lines of the Replica Calculation performed in Sec.~\ref{app:sec:replica_calculations}, one gets
\begin{align}
    \epsilon_g = \epsilon(R,b),
\end{align}
with $(R,b)$ parameters at equilibrium \textit{i.e.}~the ones that solve the saddle-point equations.

As introduced in the main manuscript, all the \textbf{generalization metrics} that we investigate in the manuscript can be expressed in terms of \textbf{True Positive Rate} (Recall, $r$) and the \textbf{True Negative Rate} (Specificity, $s$). Here we show the derivation for these two metrics:

\begin{align}
        r &= \frac{\left\llangle\left\llangle \mathbb{E}_T \left[  \left[1 - \Theta\left(  -\left(  \frac{\mathbf{w}\cdot \mathbf{S}}{\sqrt{N}} + b  \right) \left(  \frac{\mathbf{w}^0\cdot \mathbf{S}}{\sqrt{N}} + b_0  \right)\right) \right] \Theta \left(  \frac{\mathbf{w}^0\cdot \mathbf{S}}{\sqrt{N}} + b_0  \right)     \right] \right\rrangle_{\mu_{\text{train}}} \right\rrangle_{\mu_{\text{test}}} }{\rho_{test}}\\
        &= \left. \frac{1}{c_+} \int D \mathbf{S} \Theta\left(  \frac{\mathbf{w}^0\cdot \mathbf{S}}{\sqrt{N}} + b_0  \right) \left[1 - \Theta\left(  -\left(  \frac{\mathbf{w}\cdot \mathbf{S}}{\sqrt{N}} + b  \right) \left(  \frac{\mathbf{w}^0\cdot \mathbf{S}}{\sqrt{N}} + b_0  \right)\right) \right] \right\vert_{S.P.}\\
        &= \left. 1 - \frac{1}{c_+} \int D \mathbf{S} \Theta\left(  \frac{\mathbf{w}^0\cdot \mathbf{S}}{\sqrt{N}} + b_0  \right)  \Theta\left(  -\left(  \frac{\mathbf{w}\cdot \mathbf{S}}{\sqrt{N}} + b  \right) \left(  \frac{\mathbf{w}^0\cdot \mathbf{S}}{\sqrt{N}} + b_0  \right)\right) \right\vert_{S.P.} \\
        &= 1 - \left.\frac{1}{c_+} \mathrm{I}_+(R,b,b_0)\right\vert_{S.P.}\,.
    \end{align}
The derivation follows the same lines of the one for the Generalization Error. We stress that metric is evaluated at the saddle point for the order parameters,
\begin{align}
    s &= \frac{\left\llangle\left\llangle \mathbb{E}_T \left[  \left[1 - \Theta\left(  -\left(  \frac{\mathbf{w}\cdot \mathbf{S}}{\sqrt{N}} + b  \right) \left(  \frac{\mathbf{w}^0\cdot \mathbf{S}}{\sqrt{N}} + b_0  \right)\right) \right] \Theta \left( -\frac{\mathbf{w}^0\cdot \mathbf{S}}{\sqrt{N}} - b_0  \right)     \right] \right\rrangle_{\mu_{\text{train}}} \right\rrangle_{\mu_{\text{test}}} }{1-\rho_{test}}\\
    &= 1 - \left.\frac{1}{c_-} \mathrm{I}_-(R,b,b_0)\right\vert_{S.P.}\,.
\end{align}
The other metrics shown in the main paper can be inferred from $r$ and $s$ through Eqs.~(\ref{eq:r}--\ref{eq:err}) in the main paper.
Additionally, we here provide the expressions of the precision for negative samples $p(-)$, and of the negative-class F1-score, $F_1(-)$.

\begin{align}
    p(-) &= \frac{s}{s + \frac{\rhotest}{1- \rhotest} (1-r)}\,,
    \label{eq:fallout}
    \\
    F_1(-) &= 2\cdot \frac{p(-) \cdot s}{p(-)+s}\,.\label{eq:f1neg}
\end{align}

\section{WHEN THE TEACHER BIAS IS KNOWN}
\label{app:fixedb}
Here we consider the simpler scenario where the student's bias is not learned but fixed at $b = b_0$, corresponding to the situation in which the student has prior knowledge of the teacher's bias. This case is particularly insightful because it reveals some underlying symmetries of the problem and clarifies the concept of \textit{informative samples} introduced in the main text.

\begin{figure}[!ht]
    \centering
    \includegraphics[width=0.47\linewidth]{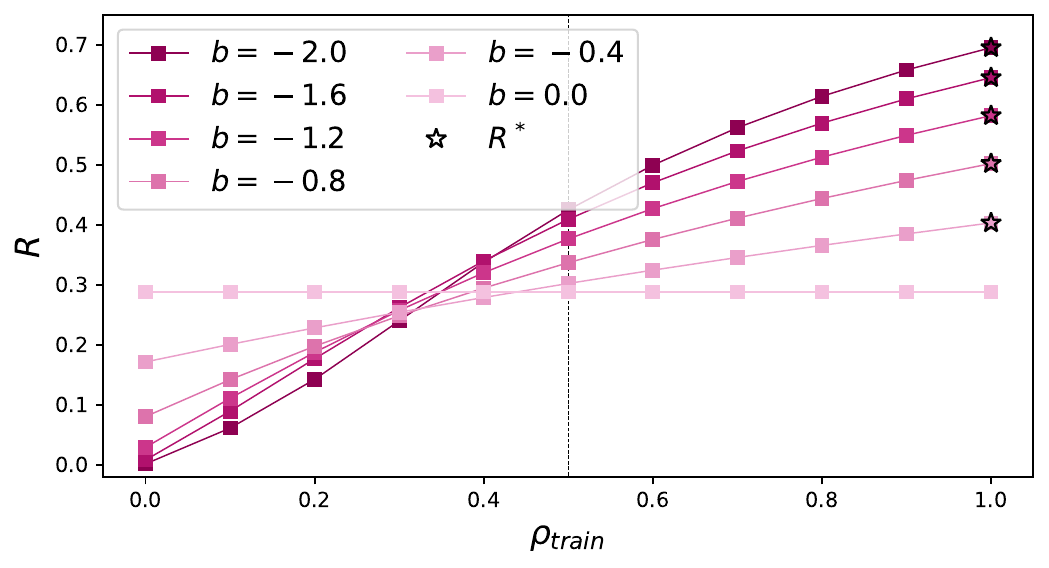}
    \includegraphics[width=0.47\linewidth]{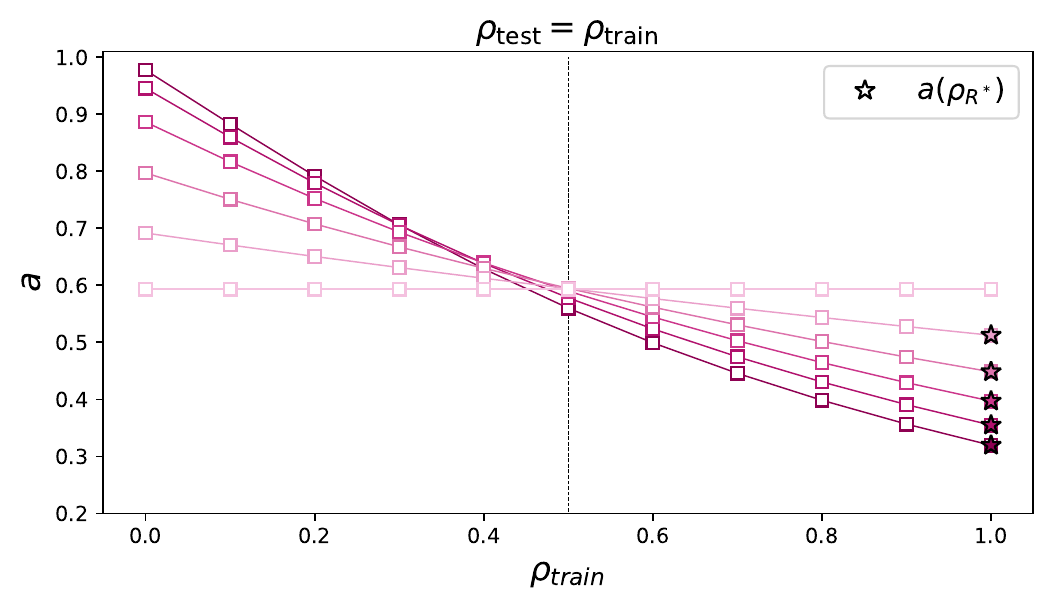}
    \caption{Overlap and accuracy on the spherical teacher-student perceptron, with the constraint $b=b_0$.
    \textbf{Left:} Teacher-student overlap $R$ as a function of $\rho_{\mathrm{train}}$, for $\alpha=0.7$ and $T=0.5$. Stars indicate the point where the overlap is maximized. The vertical line indicates $\rho_\mathrm{train}=0.5$. \textbf{Right:} Test-set accuracy $a$ with $\rhotest=\rho_{\mathrm{train}}$ as a function of $\rho_{\mathrm{train}}$, for $\alpha=0.7$ and $T=0.5$. Stars indicate the point where the overlap is maximized which correspond to a low accuracy on the test set. 
    }
    \label{fig:R_acc_fixed_b}
\end{figure}

\paragraph{Optimal training.}
In this setup, the self-consistent Eqs.~(\ref{eq:saddle1}--\ref{eq:saddle5}) presented in App.~\ref{app:sec:replica_calculations} simplify, reducing to just the first four equations, since the student's bias $b$ is fixed and does not need to be fixed self-consistently. The relevant information is thus captured entirely by the teacher-student overlap $R$. A key consequence of this fixed bias is that \textbf{translations of the student’s hyperplane are not permitted}. Therefore, the alignment between teacher and student depends solely on the density of samples near the teacher’s hyperplane, regardless of their class labels. When $b_0 \neq 0$, one class becomes more informative about the teacher's direction, meaning that samples closer to the teacher’s hyperplane provide more information about its position. This becomes particularly evident here: when $b_0 \neq 0$, one class is inherently more informative, and to maximize the overlap $R$, \textbf{the optimal training set would ideally consist solely of samples from the minority class}. This phenomenon is illustrated in Fig.~\ref{fig:R_acc_fixed_b}--(left), where the resulting $R$ from training is plotted against $ \rho_{\mathrm{train}} $ for various values of the teacher's bias $ b_0 $.
We observe that as $ |b_0| $ increases, this effect becomes more pronounced, as the minority class samples become increasingly concentrated near the teacher’s hyperplane.

\paragraph{Invariance under sample reflection.}
The case $b = b_0 = 0$ is particularly interesting because it highlights a symmetry in the problem. Here, both classes contribute equally to informativeness, meaning there is no advantage in having more samples from one class over the other. This symmetry is reflected in the flat curve in Fig.~\ref{fig:R_acc_fixed_b}--(left). It also manifests in the Free-Energy function, where the two energetic terms $ \mathcal{G}_r^\pm $, defined in Eq.~\eqref{eq:energetic_terms_separate}, become equal when $ b_\sigma = b_0 = 0 $. In fact, by applying the change of variables $ \mathbf{S} \to -\mathbf{S} $ in the integral, we recover $ \mathcal{G}_r^- $ from $ \mathcal{G}_r^+ $ and vice versa. This symmetry arises because the Boltzmann weight of each sample is identical regardless of its label, meaning that as long as the total number of samples remains fixed, the free-energy remains the same. In essence, when $ b = 0 $, there exists a bijection between flipping the labels and flipping the samples $ \mathbf{S} $. Thus, in the second integral of Eq.~\eqref{eq:need_of_b}, imposing a label flip also imposes a reflection in the data space, leading to the problem's invariance under sample reflection.

\paragraph{Test accuracy.}
Another important insight from this simplified case is that evaluating simple accuracy $ a $ on a test set with the same imbalance as the training set ($ \rhotest = \rho_{\mathrm{train}} $) can be misleading. We observe that the value of $ \rho_{\mathrm{train}} $ which maximizes accuracy often corresponds to a lower overlap $ R $. This discrepancy arises because a higher density of samples near the hyperplane increases the likelihood of misclassification, lowering accuracy even when the alignment between teacher and student is quite strong. This effect is demonstrated in Fig.~\ref{fig:R_acc_fixed_b}--(right), where the accuracy on the test set is plotted against $\rho_{train}$.

\section{THE OPTIMAL TRAIN IMBALANCE}
\label{app:rhostar}

\begin{figure}[!ht]
\centering
\includegraphics[width=0.485\textwidth]{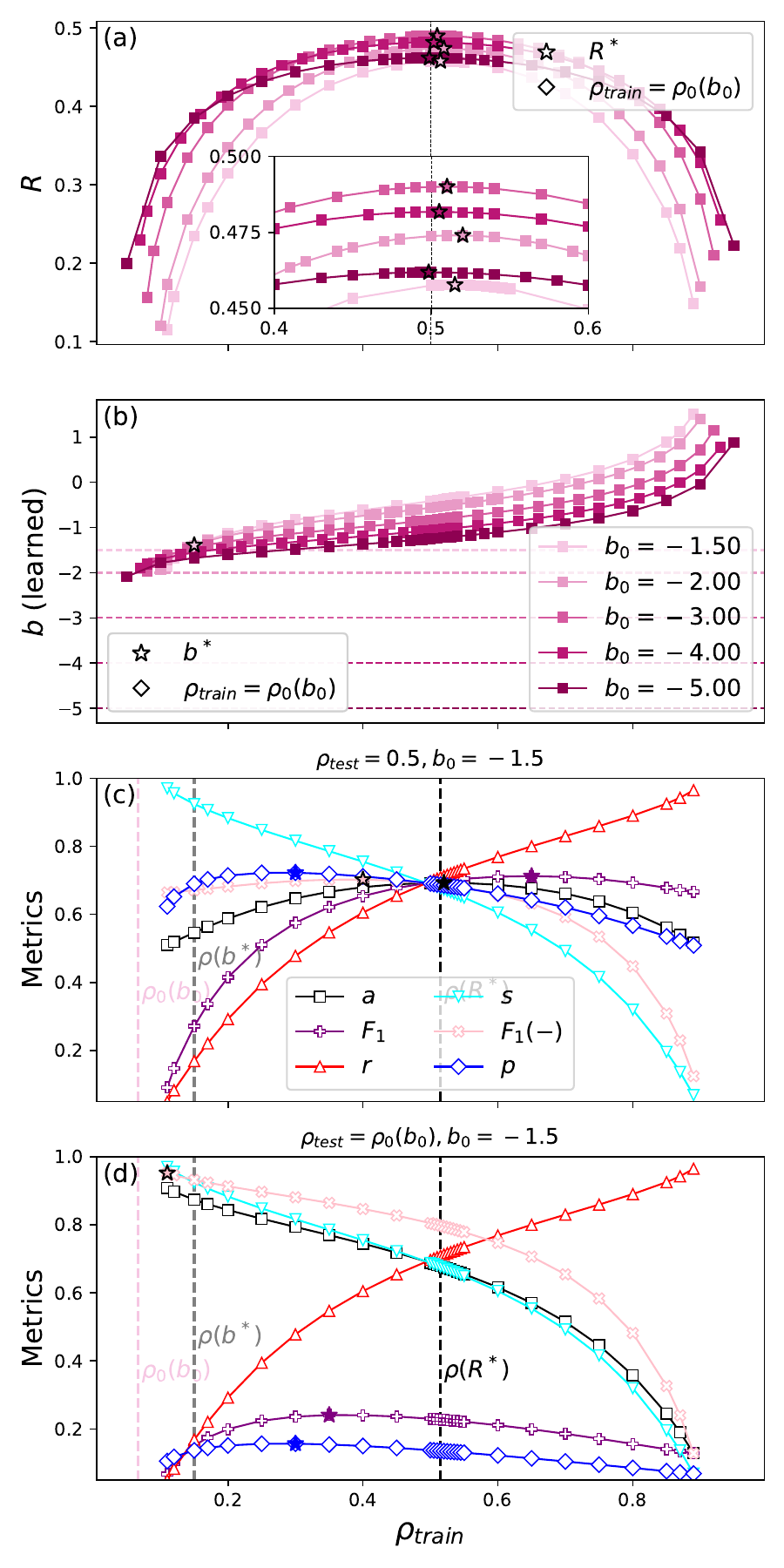}
\includegraphics[width=0.485\textwidth]{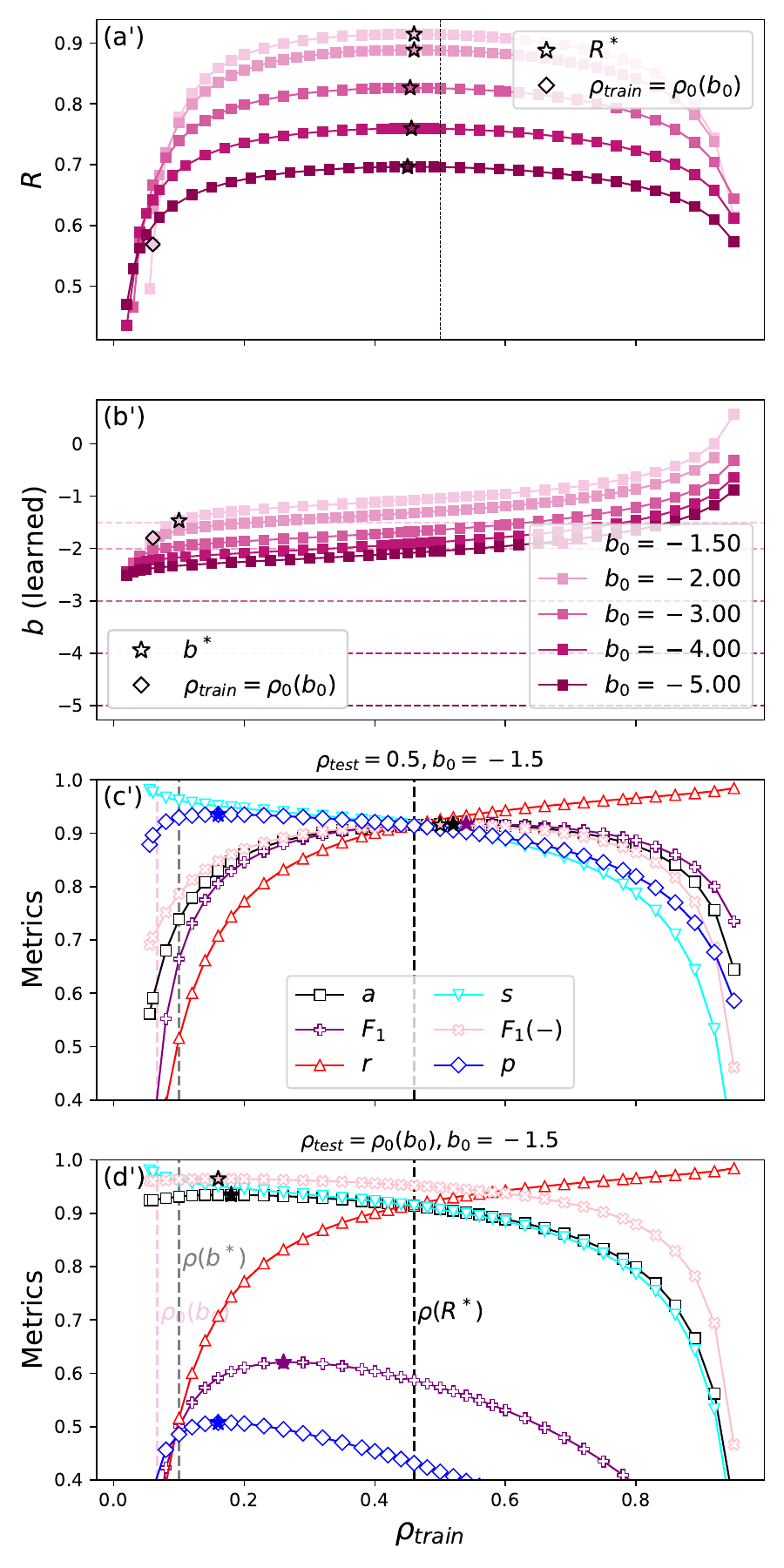}
\caption{
\textit{Performance as function of $\rhotrain$}, for $\alpha=1.1$ (left) and $\alpha=8$ (right).
Analytical results for $T=0.5$. 
The bias takes values $b_0=-1.5,-2,-3,-4,-5$, which correspond to $\rho_0(b_0)=7\cdot10^{-2}, 2\cdot10^{-2}, 6\cdot10^{-3}, 1\cdot10^{-3}, 3\cdot10^{-5}, 3\cdot10^{-7}$.
\textbf{(a):} Student overlap $R$. Stars indicate the point where the overlap is maximized. Diamonds, which as in Fig.~\ref{fig:theoretical_results_vs_rho} indicate the performance at $\rhotrain=\rho_0$, were computed only for $\alpha=8, b_0=-1.5$. The vertical line indicates $\rhotrain=0.5$. The inset is a zoom. 
\textbf{(b):} Same as (a), but for the student bias $b$. The horizontal lines indicate $b_0$. Now the stars indicate the points where bias is learnt perfectly (\ie when $b=b_0$ is reached), and the diamonds that indicate the $b$ learned under $\rhotrain=\rho_0$ (shown when computed).
\textbf{(c):} Accuracy, recall, sensitivity, precision, F1 score, and F1 score of the negative class, for $b_0=-0.6$, $\rhotest=0.5$. The stars indicate the peak of each curve. The vertical lines indicate $\rho_0$, the imbalance $\rho(b^*)$ at which the bias is optimal, and that at which the overlap is optimal, $\rho(R^*)$.
\textbf{(d):} Same as (c), but for $\rhotest=\rho_0(b_0)$. 
}
\label{app:fig:theoretical_results_vs_rho_all_metrics}
\end{figure}

In this section, we show that the results provided through Fig.~\ref{fig:theoretical_results_vs_rho} are robust to hyperparameter changes. Moreover, we show that the shift in the optimal training imbalance $\rho(R^*)$ depends non-monotonously on the hyperparameters, and that this shift can be both positive and negative (meaning that $\rho(R^*)$ can be both larger and smaller than 0.5).

Fig.~\ref{app:fig:theoretical_results_vs_rho_all_metrics} displays the same quantities as Fig.~\ref{fig:theoretical_results_vs_rho}, but shows what happens for more extreme $b_0$ values (both panels) and for a much larger $\alpha$ (right panel). As in the main text, these results are obtained by self-consistently solving the set of equations [Eqs.~(\ref{eq:saddle1}--\ref{eq:saddle5})] obtained for the order parameters,
and deriving the values of the performance metrics values, as explained in App.~\ref{app:sec:train_and_generalization_metrics}.

\paragraph{More data, better performance.}
A first obvious and unsurprising fact is that when data is more abundant ($\alpha=8$), performance is overall better for all relevant metrics ($a,p,F_1$). In the following we are interested in relative performance and trends rather than on sheer performance.

\paragraph{Position of the peak $\boldsymbol{\rho(R^*)}$.}
The most important and striking point is that for $\alpha=8$  (panel (a')), the $\rhotrain$ at which $R$ peaks ($\rho(R^*)$) is clearly away from $0.5$, around $0.45$, confirming that the optimal $\rhotrain$ is in general not $0.5$, can be far away from it, and that it can be either larger or smaller than this commonly used value.
We also note (panel (c')) that while $\rho(a^*)$ is also away from $0.5$, it is around $0.55$, \ie on the other side of the $\rhotrain=0.5$ reference. This indicates that, although we have identified the balanced accuracy as the metric which best reproduces the overlap, there is still a qualitative divide between $a_\mathrm{bal}$ and $R$.

\paragraph{Strong biases.}
In Fig.~\ref{fig:theoretical_results_vs_rho}(b), the  learned bias $b$ could be smaller or larger than the teacher's bias $b_0$, depending on $\rhotrain$. 
Here (Fig.~\ref{app:fig:theoretical_results_vs_rho_all_metrics}(b,b')), under strong intrinsic imbalance (extreme $\rho_0$ values), we note that the bias learnt is essentially always under-estimated (in absolute value) by the student; when it is not the case, performance is extremely bad (for all metrics).
This is consistent with our argument about the region $\mathcal{R}$ being quite empty ($\mathcal{R}$ is the region between the positive samples -- squished against the teacher's hyperplane -- and the typical location of the negative samples). As $|b_0|$ grows, the positive samples are increasingly squished against the teacher's hyperplane while the negative samples remain around the origin. Thus the region $\mathcal{R}$ grows in size and it is thus energetically favorable for the student to locate the hyperplane in this region, as it is easy to get $0$ error there, even if the alignment $R$ is not perfect.

\begin{figure}[!ht]
    \centering
    \includegraphics[width=0.31\linewidth]{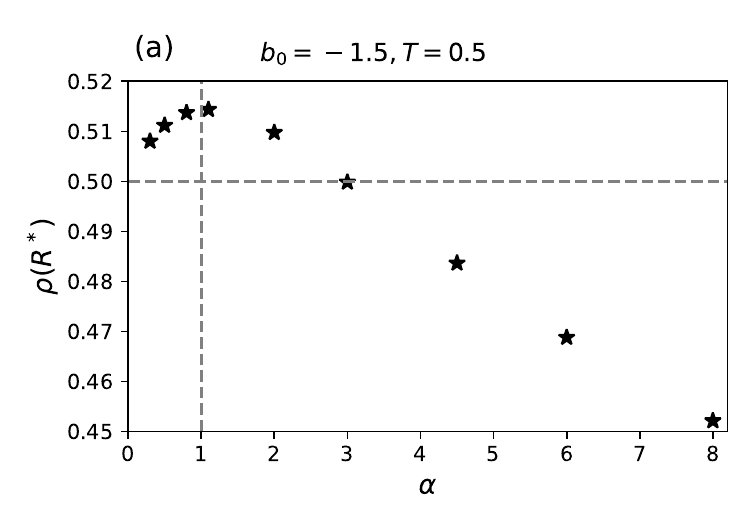}
    \includegraphics[width=0.32\linewidth]{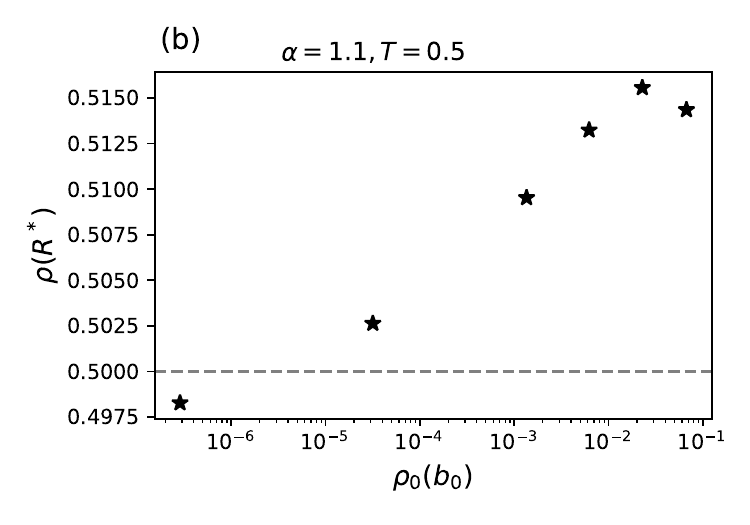}
    \includegraphics[width=0.32\linewidth]{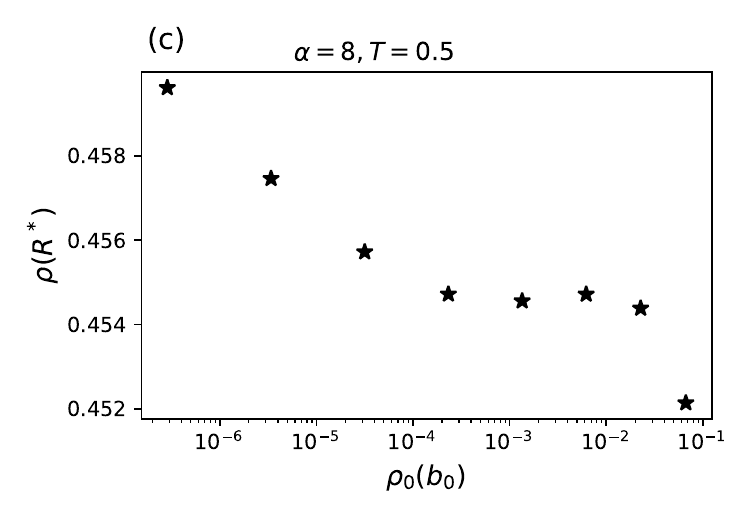}
    \caption{
    \textit{Optimal $\rhotrain$ as function of the control parameters $\alpha$ and $\rho_0(b_0)$}. \textbf{
    (a):} $\rho(R^*)$ as function of data abundance $\alpha$.
    \textbf{(b):} $\rho(R^*)$  as function of the intrinsic imbalance $\rho_0$ (controlled by $b_0$), for $\alpha=1.1$.
    \textbf{(c):} Same as (b), for $\alpha=8$.
    Dashed grey lines highlight the values $\rhotrain=0.5$ or $\alpha=1$.
    }
    \label{app:fig:rhostar_vs_alpha}
\end{figure}

\paragraph{Non-monotonicity of $\boldsymbol{\rhotrain(R^*)}$.}
As mentioned in the main text, $\rhotrain(R^*)$ does not need to be a monotonic function of $b_0$, nor of $\alpha$.
This can be seen more clearly in Fig.~\ref{app:fig:rhostar_vs_alpha}.
In panel (a), we see that $\rhotrain(R^*)$ is a non-monotonic function of $\alpha$.
In panels (b,c) of Fig.~\ref{app:fig:rhostar_vs_alpha} we note that the dependency in $\rho_0$ (itself controlled monotonously by $b_0$) is non trivial and itself depends on $\alpha$.
This is challenging to interpret in simple terms, as one could expect a priori that more intrinsic imbalance would have a straightforward effect on $\rhotrain$.

\begin{figure}[!ht]
    \centering
    \includegraphics[width=0.31\linewidth]{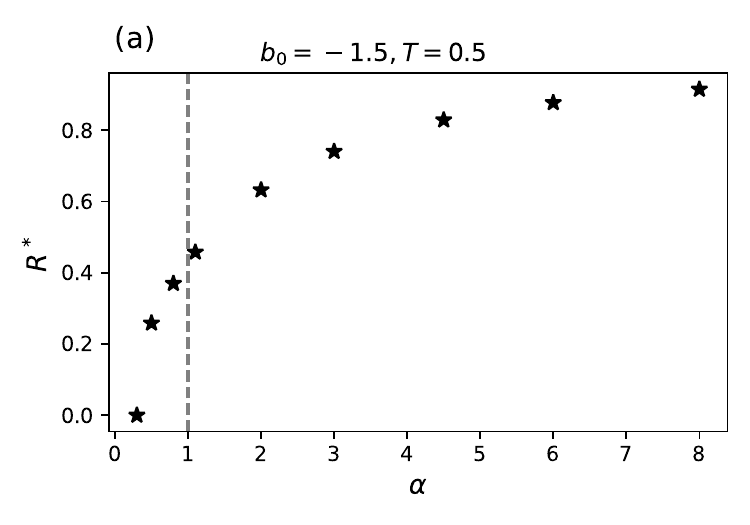}
    \includegraphics[width=0.32\linewidth]{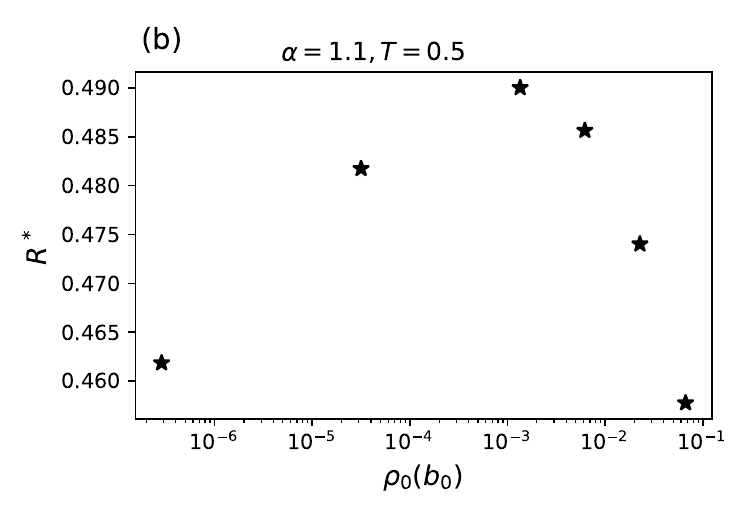}
    \includegraphics[width=0.32\linewidth]{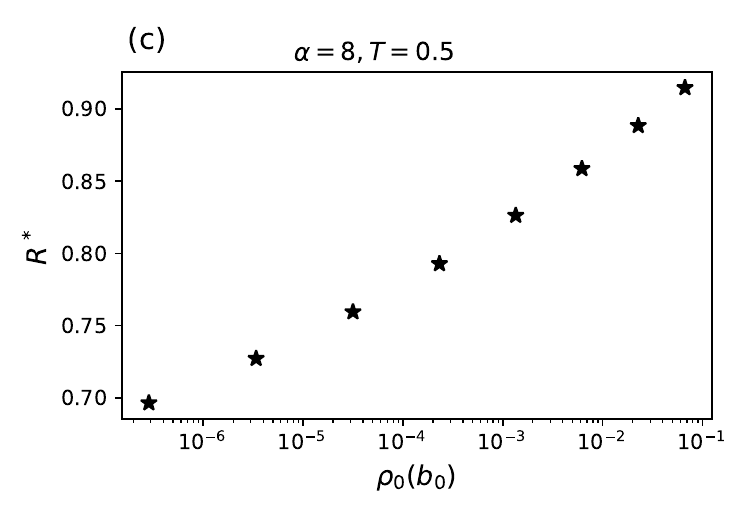}
    \caption{
        \textit{Optimal overlap, $R^*$, found by fine-tuning $\rhotrain$, as function of the control parameters $\alpha$ and $\rho_0(b_0)$}
     \textbf{(a):} $R^*$ as function of data abundance $\alpha$.
    \textbf{(b):} $R^*$  as function of the intrinsic imbalance $\rho_0$ (controlled by $b_0$), for $\alpha=1.1$.
    \textbf{(c):} Same as (b), for $\alpha=8$.
    }
    \label{app:fig:Rstar_vs_alpha}
\end{figure}

\paragraph{Trend of $\boldsymbol{R^*(b_0)}$.}
In Fig.~\ref{app:fig:Rstar_vs_alpha}(a), we show that the optimal reconstruction of the teacher ($R^*$) is an increasing function of the amount of data $\alpha$.
In Fig.~\ref{app:fig:Rstar_vs_alpha}(b,c), we show that $R^*$ can be non-monotonous, especially in the case $\alpha=1.1$.

\paragraph{Relative importance of intrinsic and train imbalance.}
In Fig.~\ref{app:fig:theoretical_results_vs_rho_all_metrics}(a) we note that $R$ mildly depends on $\rho_0$ but strongly depends on $\rhotrain$.
This trend is reversed when $\alpha=8$ (Fig.~\ref{app:fig:theoretical_results_vs_rho_all_metrics}(e)): $R(\rhotrain)$ curves are relatively more flat and spaced between them, indicating a strong dependence on $\rho_0$ and relatively weaker one on $\rhotrain$.
This could suggest that, if one has access to only small amounts of data, the choice of $\rhotrain$ would become crucial. On the contrary, with large amounts of data,
$\rho_0$ plays a more relevant role than $\rhotrain$. Our exploration of the hyperparameter space is however not extensive enough to confirm this conjecture.




\begin{figure}[!ht]
    \centering
    \includegraphics[width=0.47\linewidth]{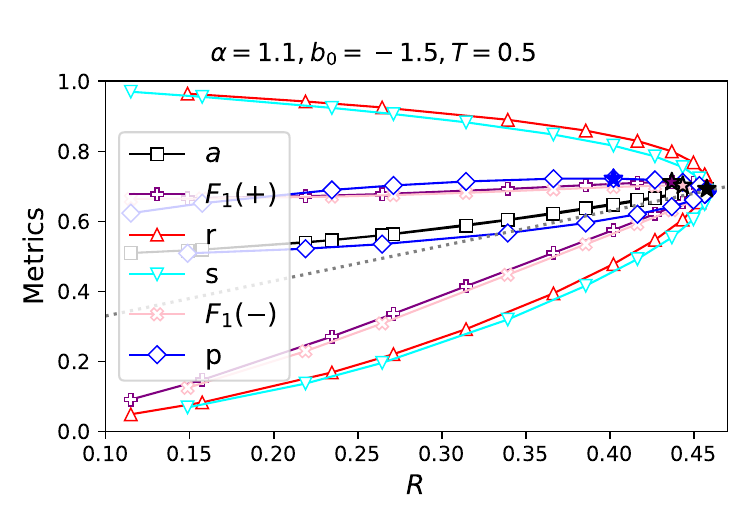}
\includegraphics[width=0.47\linewidth]{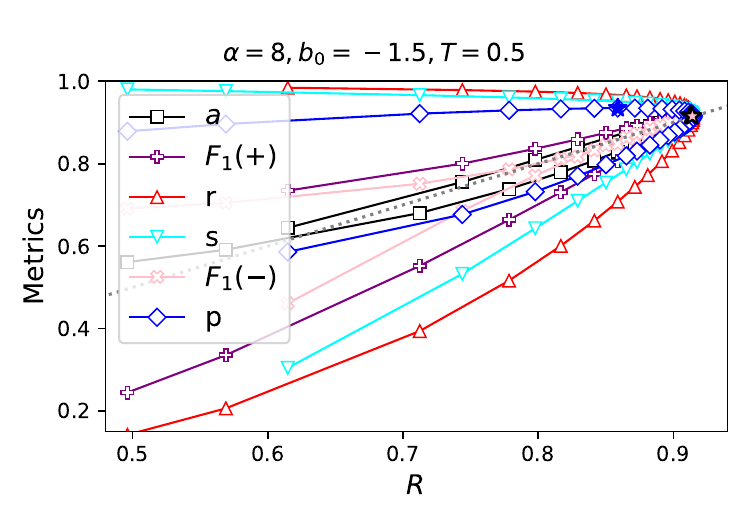}
\includegraphics[width=0.47\linewidth]{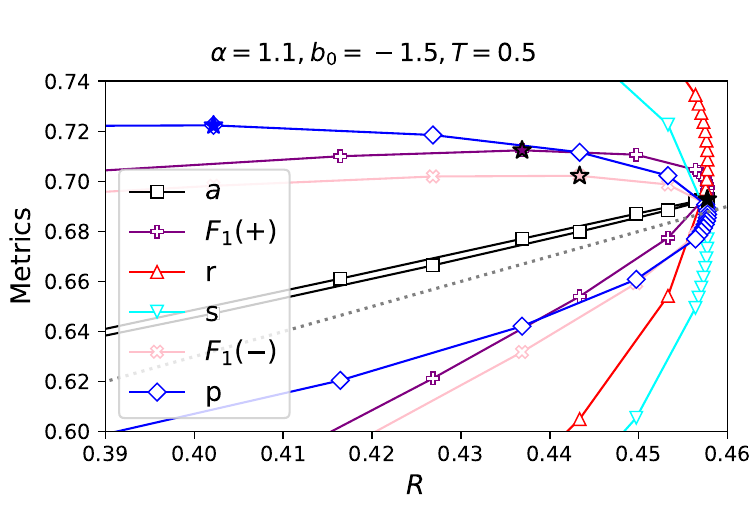}
\includegraphics[width=0.47\linewidth]{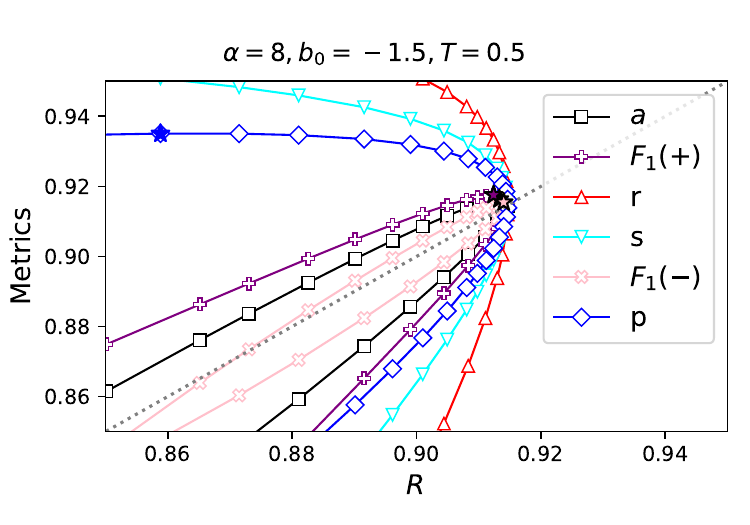}
    \caption{
    \textit{Metrics as a function of $R$,} for $\rhotest=0.5$.
    The stars indicate the maximum value of each curve when it's non trivial.
    \textbf{Left:} $\alpha=1.1$.
    \textbf{Right:} $\alpha=8$.
    The bottom figures are zooms of the top figures.  The grey dotted lines are parallel to the diagonal $y=x$ line, as guide to the eye.
    }
    \label{app:fig:metrics_vs_R}
\end{figure}

\paragraph{Correlation of Metrics with $\boldsymbol{R}$}
In Fig.~\ref{app:fig:metrics_vs_R}, we show how each metric correlates with the overlap, when tuning $\rhotrain$, and with $\rhotest=0.5$.

The best metric is identified through two indicators:
\begin{enumerate}
    \item how close the rightmost point (maximal overlap $R$, what we aim for) is to the topmost point (maximal metric, what we can infer, highlighted with a star).

    \item
    how much the metric is almost in bijection with the overlap $R$, 
    so that when the peak of the metric is passed, the optimal overlap is passed too. 
\end{enumerate}

From Fig.~\ref{app:fig:metrics_vs_R},
one can see that no metric is perfectly representative of $R$, and that the balanced accuracy  $a_\mathrm{bal}$ (the curves are for $\rhotest=0.5$, so  $a=a_\mathrm{bal}$) is the one that correlates best, since the two branches of the curve are close together and well-aligned to a 1:1 constant slope. This is particularly clear in the bottom-right plot, where we see that $a_\mathrm{bal}$ peaks most to the right.

We stress that these results indicate which is the best metric to identify which $\rhotrain$ should be used, and not whether the metric is a good metric for testing.

\begin{figure}[!ht]
    \centering
    \includegraphics[width=0.47\linewidth]{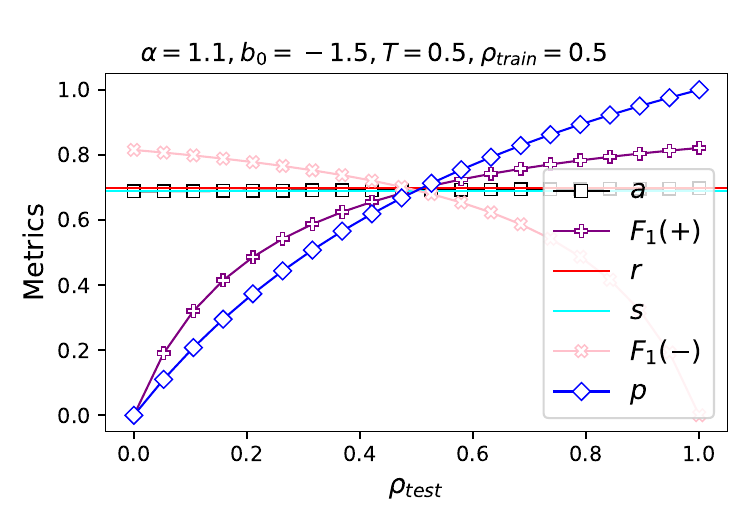}
\includegraphics[width=0.47\linewidth]{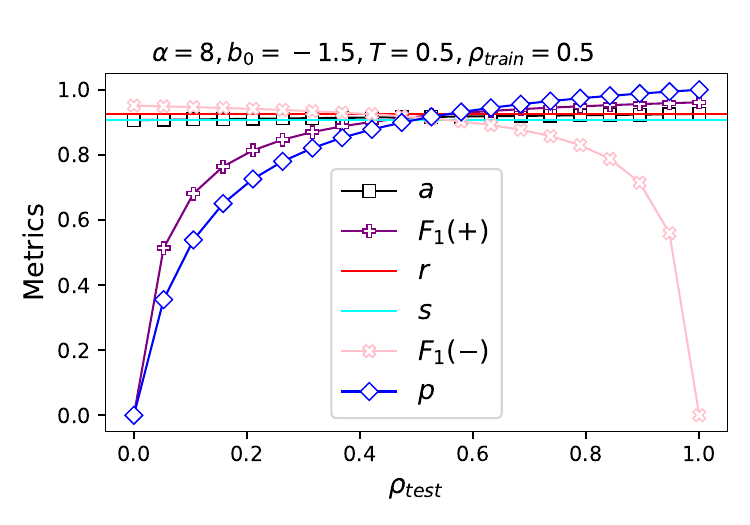}
    \caption{
    \textit{Metrics as a function of $\rhotest$}, for $\rhotrain=0.5$.
    }
    \label{app:fig:Metrics_vs_rhotest}
\end{figure}

\paragraph{Metrics Dependence on $\boldsymbol{\rhotest}$}
In Fig.~\ref{app:fig:Metrics_vs_rhotest} we show how metrics depend on $\rhotest$, as this is a more usual setup, since the $\rhotest$ can easily be changed (while tuning $\rhotrain$ implies re-training the model for each new value of $\rhotrain$). 
An interesting point is that the accuracy $a$ (not balanced since $\rhotest$ is varying), which is a weighted sum of recall $r$ and specificity $s$ which here have very similar values, is very robust against changes in $\rhotest$ (although not formally constant), for both $\alpha$ values.
This is a result of using $\rhotrain \approx 0.5$: for different values, this feature is lost.   
However one is usually interested in getting the best performance (higher $a$ or other Metric at some $\rhotest$), not in getting equal $r$ and $s$.

\section{EXPERIMENTS}
\label{app:experiments}

\rev{We consider three experimental setups, which allow to compare our theoretical model with more realistic setups. 
More precisely, our experiments allow to answer several questions, respectively: 
(i) what is the influence of the learning dynamics and of the read-out (activation) function on our results, within a controlled scenario that mirrors theoretical computations; 
(ii) what are the effects of dataset characteristics and model choice in a scenario more akin to practical machine learning applications; 
(iii) how realistic is the ADI assumption on data in yet more realistic AD tasks, and whether qualitative results are robust to this change of input data.}

\subsection{Perceptron Teacher Student on Gaussian data}
In this setup, we use a Spherical Perceptron within a Teacher-Student framework. The model is linear, with the weights' norm constrained to be $O(1)$ and initialized according to a normal distribution; a sigmoid function is chosen for the activation. The model is trained through the minimization of the L2 loss. This configuration closely mirrors the theoretical setup, with two key distinctions: the use of a sigmoid activation function that outputs continuous values within the $[0,1]$ range, and SGD learning dynamics, similar to real machine learning practices. It is important to note that the choice of a continuous activation function is essential to enable gradient descent dynamics, as a discontinuous function, such as the sign function $g$ in Eq.~\ref{eq:gell}, would result in gradients that are almost always  zero. 
To produce binary labels, teacher outputs are discretized, assigning a label of $1$ for values above the threshold $\mathrm{thr} = 0.5$ and $0$ otherwise. The SGD dynamic shares some characteristics with the Langevin dynamics used in the theoretical derivation, as both implement gradient descent on the training loss with added stochastic noise. The key difference lies in the correlated nature of the SGD noise, which arises from the repeated use of the mini-batch gradient estimates.

To reasonably approximate the conditions described in our theory, we set the data dimension to $N=5000$. We observe that, by further increasing $N$, our results remain stable, suggesting that this setting is close enough to the $N\to\infty$ limit. We train the student model by performing multiple passes on the whole training set (epochs) until the training loss has converged. This corresponds to the "end of training" (equilibrium) regime assumed in theoretical computations. The noise level in SGD is governed by the learning rate ($\mathrm{lr}$) and batch size ($\mathrm{BS}$), which can be approximately related to the temperature introduced in the main text as $T \sim \mathrm{lr}/\mathrm{BS}$~\cite{jastrzkebski:17}.

\begin{figure}[!ht]
\centering
\includegraphics[width=0.45\textwidth]{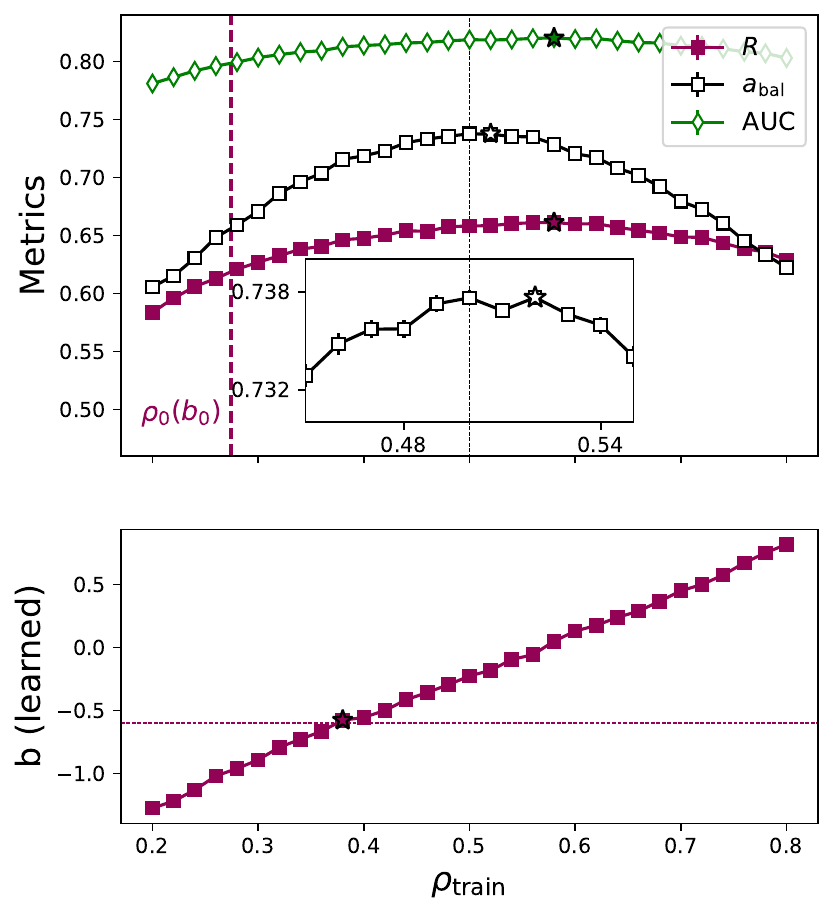}
\includegraphics[width=0.457\textwidth]{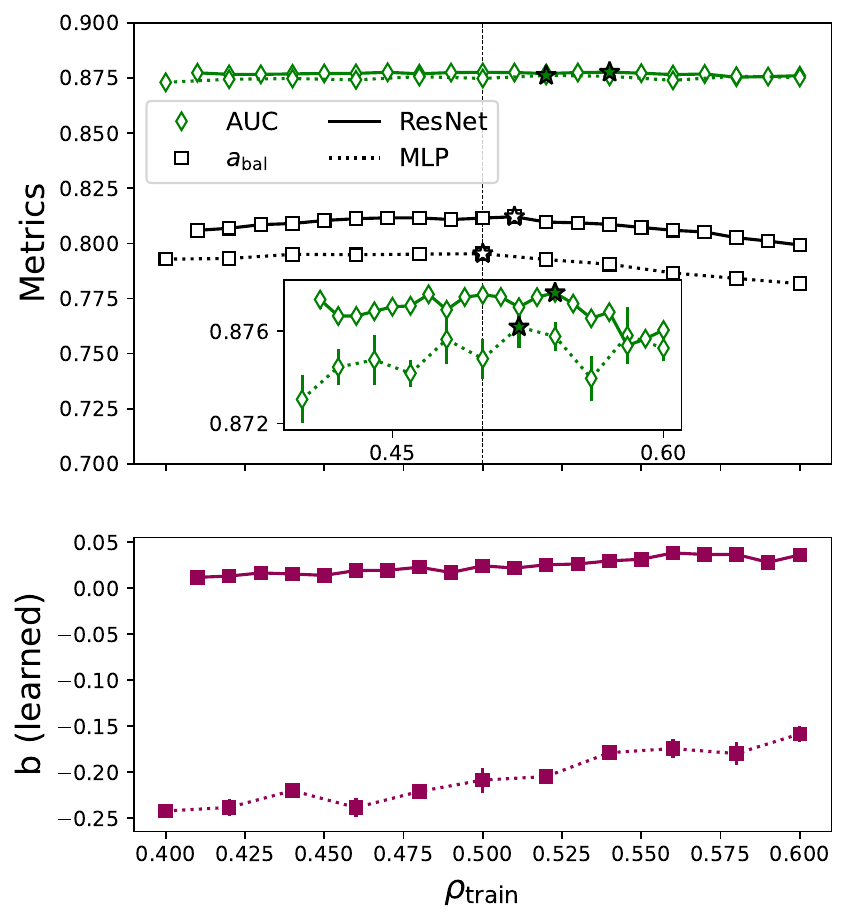}
\caption{\textbf{Left:} Perceptron TS. $b_0 = -0.6$, $\alpha=2.0$. Effective temperature $T=\frac{\text{lr}}{\text{BS}}=\frac{0.5}{20} = 2.5 \cdot 10^{-2}$. Each point represents the average over 40 re-samplings of the data and the error-bar represents its relative standard error. \textbf{Right:} MLP and ResNet34 on AD CIFAR-10. SGD optimizer, with momentum $=0.02$ and weight decay $0.01$. Each point represents the average over 10 re-samplings of the data and the error-bar represents its relative standard error. }
\label{fig:experiments_vsrho}
\end{figure}

\begin{figure}[!ht]
\centering
\includegraphics[width=0.8\textwidth]{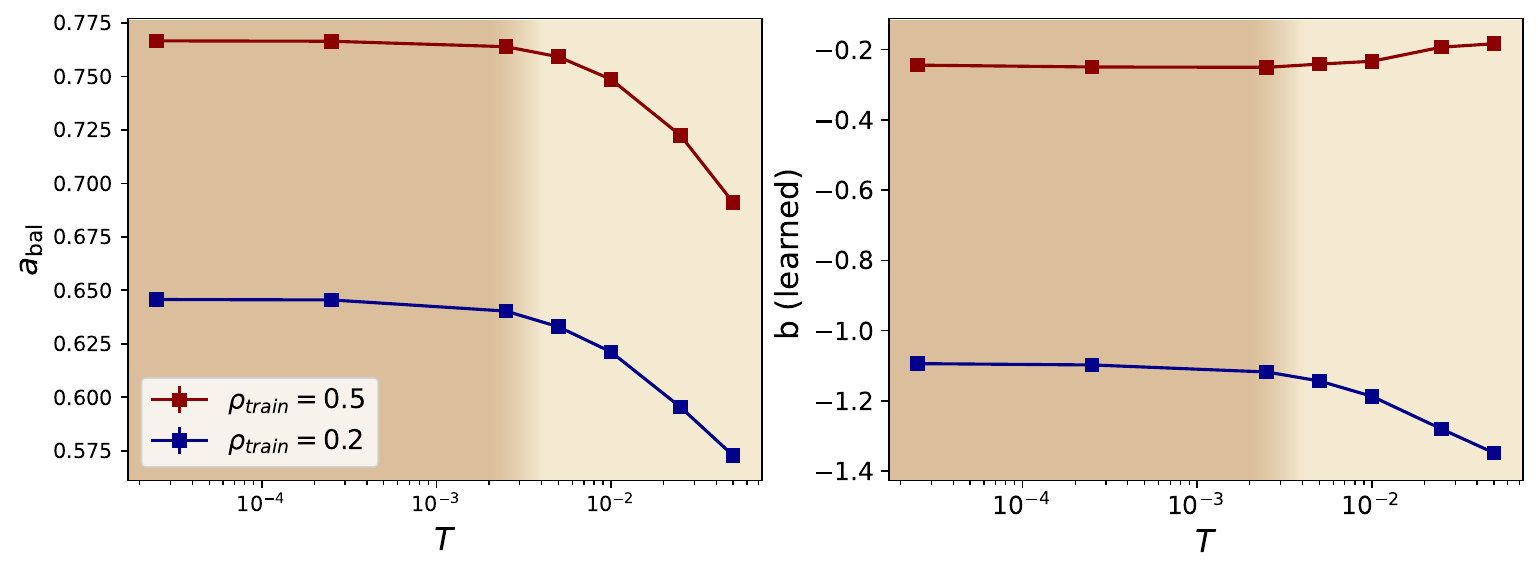}
\caption{Perceptron TS vs $T$. Temperature is varied in SGD experiments by tweaking the mini-batch size. The learning rate is fixed to be $\text{lr} = 0.05$ and the mini-batch size varies $BS = \{2000,200,20,10,5,2,1\}$.}
\label{fig:experiments_perceptronTS_BA_vsT}
\end{figure}

For each combination of control parameters $(b_0, \rho_{\mathrm{train}}, T, \alpha)$, we perform multiple runs, resampling both the dataset and teacher weights to compute the quenched average over the data distribution. Results on a balanced test set ($\rhotest = 0.5$) are illustrated in Fig.~\ref{fig:experiments_vsrho}--(left). We observe trends that are qualitatively compatible with theoretical results, and most importantly, we find a non-trivial maximum of the metrics at $\rho^* \neq 0.5$. In experiments, we also evaluate the AUC metric since a threshold is needed to discretize the output of the learned perceptron. We observe that it is rather insensitive to imbalance, confirming the findings of \cite{loffredo_restoring_2024}. Figure~\ref{fig:experiments_perceptronTS_BA_vsT} reports the trends of balanced accuracy and student's bias versus effective temperature. The experimental trends align qualitatively with the theoretical ones, identifying the low-noise and high-noise regions separated by $T^*$ as introduced in Sec.~\ref{subsec:th_results}.

\subsection{MLP and ResNet on AD CIFAR-10}
In this setup, we employ a real-world anomaly detection dataset: Anomaly Detection CIFAR-10, a standard benchmark for anomaly detection tasks (see \textit{e.g.}~\cite{Tal_PANDA_2021}). This dataset involves re-labeling the original CIFAR-10 classes such that the first class (Airplanes) is designated as the anomaly (label $+1$), while all other classes are labeled as normal samples (label $0$). CIFAR-10 consists of 60,000 samples, with 6,000 samples per class, structured and non-independent by design. Defining one class as the anomaly sets the intrinsic imbalance to $\rho_0 = 0.1$. To explore various values of $\rho_{\mathrm{train}}$, we perform sub-sampling on the dataset according to the desired level of imbalance, keeping the total number of training samples fixed at $N_{\mathrm{train}} = 6000$.

We evaluate two representative models: an MLP with one hidden layer of $16$ neurons, and an imageNet pre-trained ResNet34~\cite{kaiming_ResNet_2015} in which only the final linear layer is fine-tuned on the anomaly detection task, while all other layers remain frozen. Training is conducted via L2 loss minimization, using SGD as a learning dynamics.

In this experimental setup, not all theoretical hyper-parameters can be controlled. For instance, $\rho_0$ is fixed by the dataset, and defining $b_0$ is not feasible. Additionally, tuning $\alpha$ is challenging for both theoretical and practical reasons. Theoretically, the definition of $\alpha$ differs significantly from that in our analytical model, as the data dimensionality and model parameter count are no longer in one-to-one correspondence. Practically, varying $N_{\mathrm{train}}$ or adjusting the MLP’s hidden layer size is constrained by limited data availability and the risk of overfitting. Nonetheless, for each parameter configuration, we re-train the models multiple times, re-sampling data from the original CIFAR-10.

Results on a balanced test set, shown in Fig.~\ref{fig:experiments_vsrho}--(right), reveal a phenomenology qualitatively consistent with theoretical predictions. 
However, the effect strength and available statistics limit the conclusiveness of these findings.

\subsection{Pretrained ResNet with PCA on BTAD and MVTec}

\subsubsection{Data distribution}
\label{app:dataDistributionisADI}

\rev{In our study the input data is assumed to be Gaussian with i.i.d. components, and most importantly we assume to be in the ADI setup, \textit{i.e.}~we assume the distribution of the data of both classes to be very similar (and not split in two distinct groups, as in the MGI setup). This assumption may be questioned.}

\begin{figure}[!ht]
\centering
\includegraphics[width=0.6\textwidth]{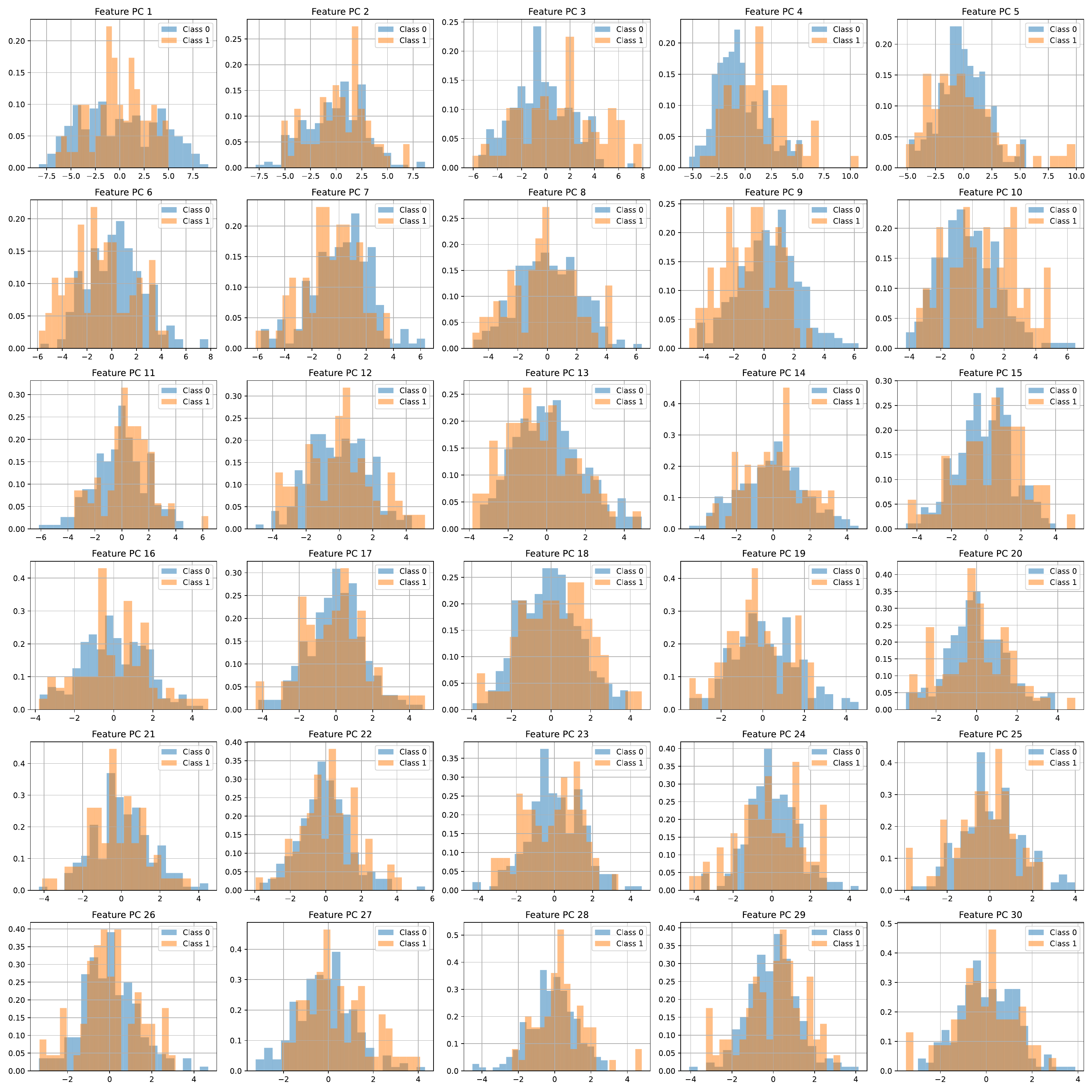}
\caption{MVTec data distribution, after a pretrained ResNet50 and a PCA.}
\label{fig:MVTEC_after_ResNet_PCA_dsitribution}
\end{figure}

\begin{figure}[!ht]
\centering
\includegraphics[width=0.6\textwidth]{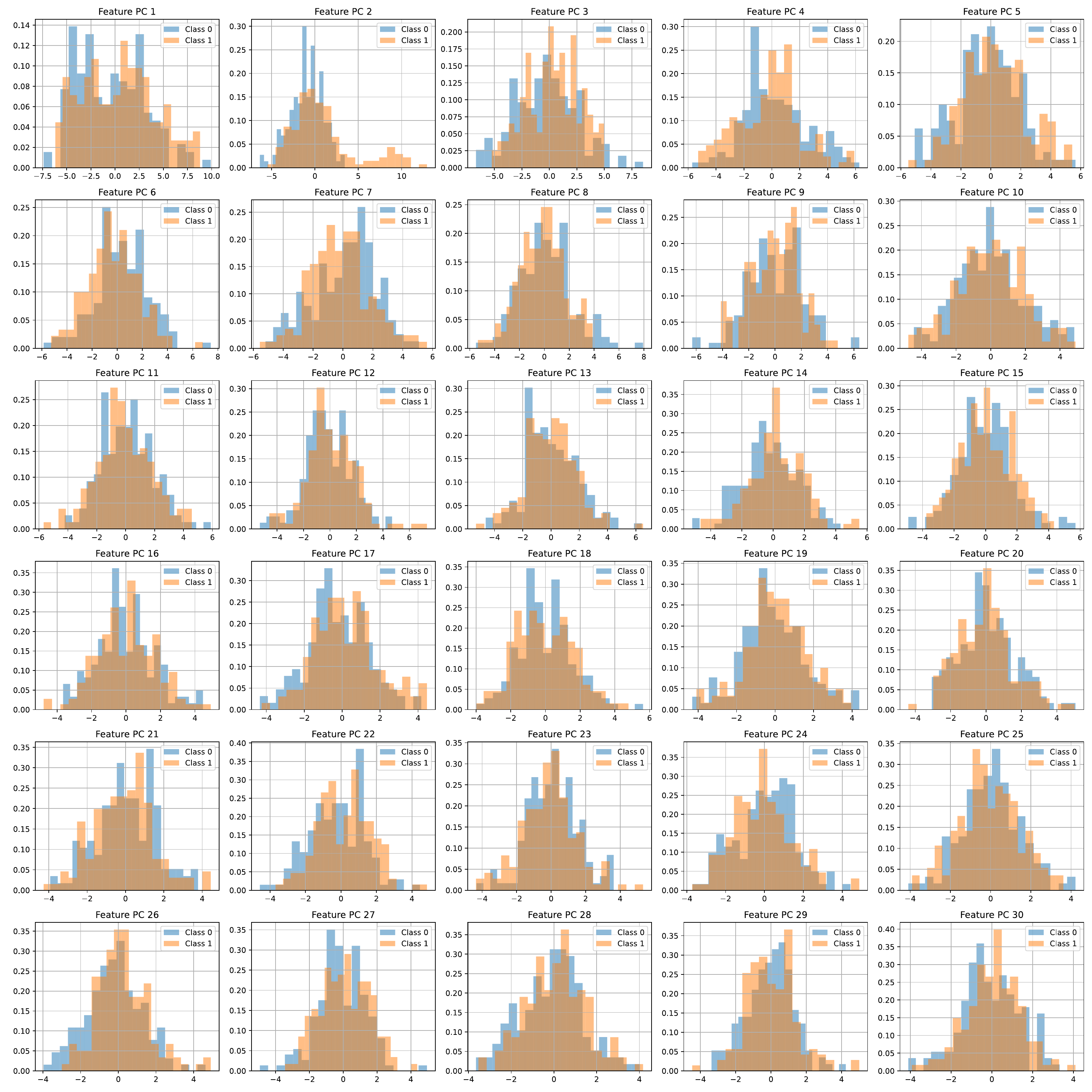}
\caption{BTAD data distribution, after a pretrained ResNet50 and a PCA.}
\label{fig:BTAD_after_ResNet_PCA_dsitribution}
\end{figure}

\rev{Here we show the relevance of this hypothesis by considering two real-world images datasets corresponding to anomaly detection problems (MVTec \cite{bergmann_mvtec_2019} and BTAD \cite{mishra2021}).
We use an ImageNet-pretrained ResNet50 as feature map and observe  the distribution of the first 30 Principal Components, conditioned on the class (see Fig.~\ref{fig:MVTEC_after_ResNet_PCA_dsitribution} and Fig.~\ref{fig:BTAD_after_ResNet_PCA_dsitribution}). 
We note that most features have a strongly overlapping distribution, \textit{i.e.}~the distribution between classes is very similar. This is indeed the case we are addressing in our modeling. 
In conclusion, for these datasets the ADI modeling is much more appropriate than the MGI setup.}

\subsubsection{Dependence on \texorpdfstring{$\rhotrain$}{the training imbalance}}

\rev{Here we further investigate the impact of using realistic data on our results, by measuring again the changes in performance as a function of $\rhotrain$.}

\begin{figure}[!ht]
\centering
\includegraphics[width=0.45\textwidth]{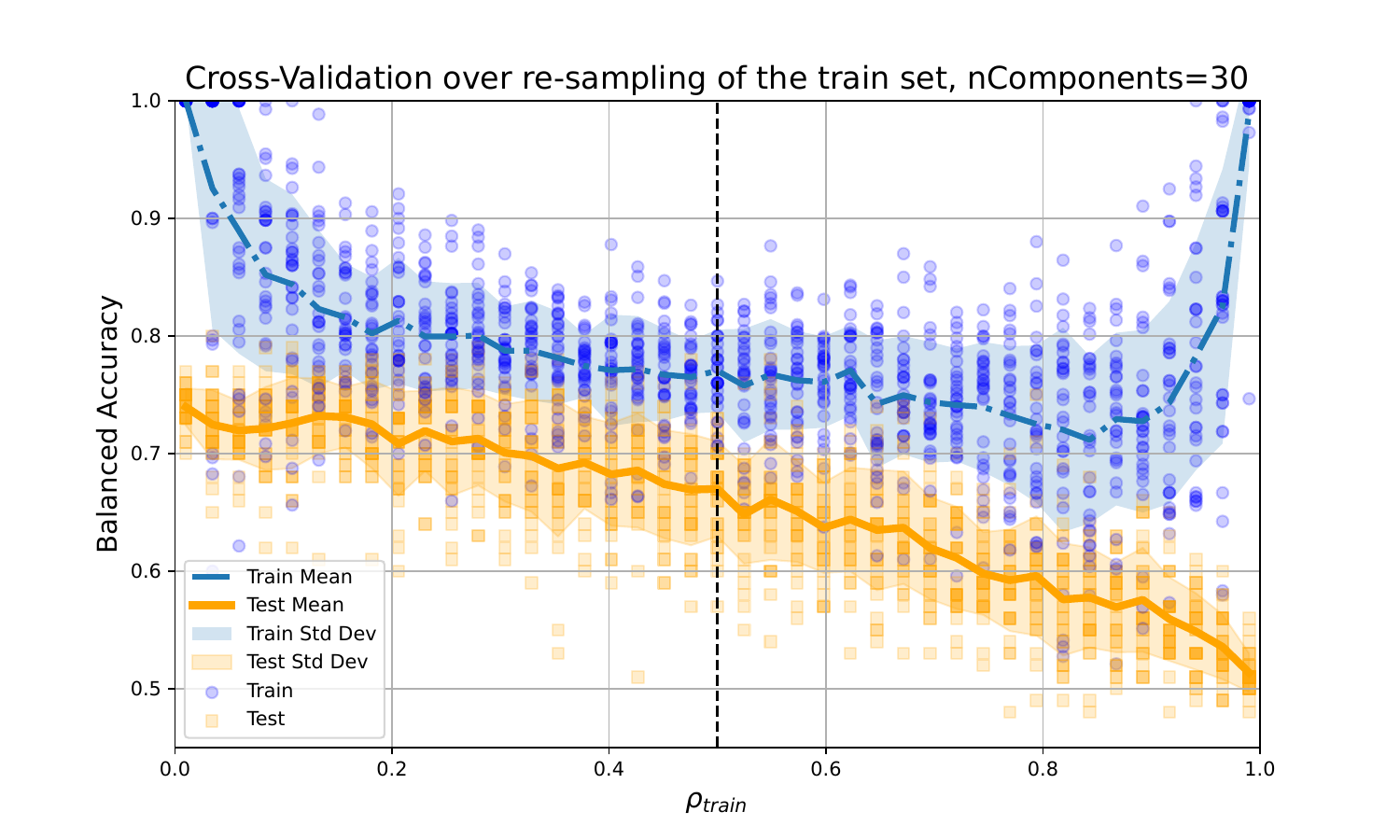}
\includegraphics[width=0.45\textwidth]{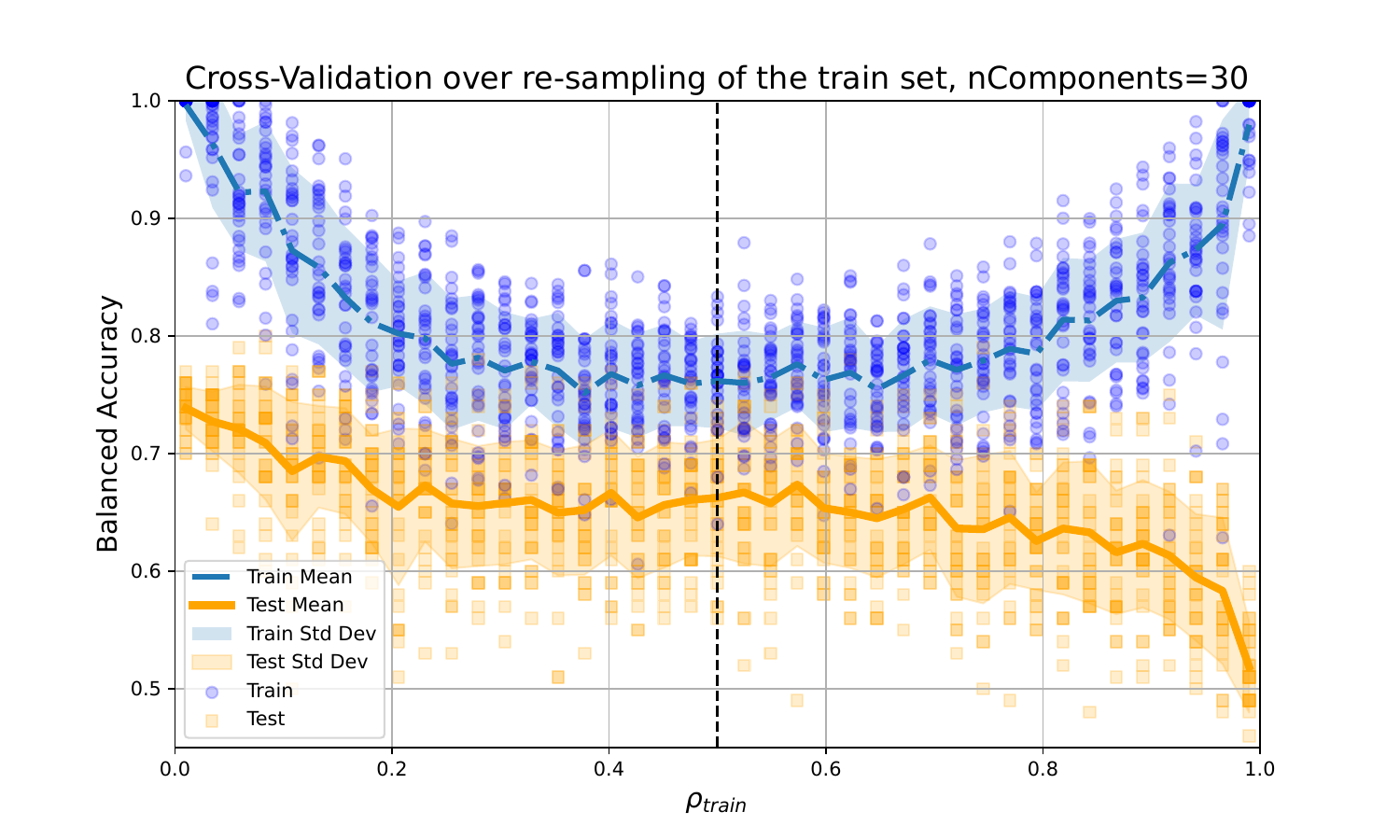}
\caption{Balanced Accuracy as a function of $\rhotrain$ for a realistic AD task. 
\\(Left): no re-balancing of the loss. The $\rhotrain$ values that strike a good balance between good test performance and moderate overfitting are in the range $\rhotrain\in [0.1,0.3]$.
\\(Right): re-balancing of the loss to correct for the effect of $\rhotrain\neq 0.5$. Balanced accuracy is lower (in the relevant range).
}
\label{fig:BTAD_after_ResNet_PCA_Perceptron}
\end{figure}

\rev{For the BATD dataset \cite{mishra2021}, we again use an imageNet-pretrained ResNet50 as backbone (with 2048 features) followed by a PCA down to 30 components.
We can then train a Perceptron model (single layer, similar to our theoretical model) and vary $\rho_{train}$ from 0 to 1. 
As we have little train and test data ($N_{train}=300$ with $250$ normal samples and $50$ anomalous ones, and $N_{test}=100$ with balanced classes), for each $\rhotrain$ we shuffle the train data 30 times, and sub-sample from this shuffle, to obtain the various $\rho_{train}$ we are interested in (keeping the total $N_{train}$ fixed).  
We average over these 30 re-samplings to estimate a mean performance (train and test balanced accuracies).
In Fig.~\ref{fig:BTAD_after_ResNet_PCA_Perceptron}~(left),
it is clear that the optimal $\rho_{train}$ is not 0.5, but rather in the range $\rhotrain^*\in [0.1,0.3]$ (searching for a good test accuracy while keeping overfitting moderate). The key point for us is that $\rhotrain^*$ is non trivial (it is not 0.5).}

\rev{As an aside, we note in Fig.~\ref{fig:BTAD_after_ResNet_PCA_Perceptron}~(right) that using re-weighting of the loss to correct for the effect of $\rhotrain\neq 0.5$ is actually detrimental (in terms of test accuracy, all other things being equal).
To be precise, except for the (very overfitting, hard-to-trust) point at $\rhotrain=0.01$ (corresponding to a single example in the anomaly class), and for overly large values of $\rhotrain$ (where performance is poor anyways), in this case it is always better to not re-weight the classes.}

\end{document}